\documentclass[default,iicol]{sn-jnl}

\jyear{2021}%

\theoremstyle{thmstyleone}%
\theoremstyle{thmstyletwo}%

\theoremstyle{thmstylethree}%

\definecolor{mygray}{gray}{.94}
\newcommand{\textBC}[2]{\textbf{\textcolor{#1}{#2}}}

\usepackage{color, colortbl}
\usepackage{xcolor}
\usepackage{url}
\usepackage{hyperref}
\hypersetup{
    colorlinks=true,
    linkcolor=blue,
    urlcolor=blue,
}
\definecolor{ggray}{RGB}{127,127,127}
\definecolor{reda}{RGB}{202,0,0}
\definecolor{mycellreda}{RGB}{202,0,0}
\definecolor{redb}{RGB}{217,148,143}
\definecolor{myyellow}{RGB}{190,144,0}
\definecolor{mygreen}{RGB}{0,136,51}
\definecolor{myblue}{RGB}{0,102,204}
\definecolor{mycellblue}{RGB}{0,102,204}

\begin{document}

\title[OOD-MAE]{\textbf{M$^{2}$SNet: Multi-scale in  Multi-scale Subtraction Network for Medical  Image  Segmentation}}

\author[1,3]{\fnm{Xiaoqi} \sur{Zhao}}
\author[1]{\fnm{Hongpeng} \sur{Jia}}
\author[1,4]{\fnm{Youwei} \sur{Pang}}
\author[2]{\fnm{Long} \sur{Lv}}
\author[2]{\fnm{Feng} \sur{Tian}}
\author[1]{\fnm{Lihe} \sur{Zhang}}
\author[2]{\fnm{Weibing} \sur{Sun}}
\author[1]{\fnm{Huchuan} \sur{Lu}}

\affil[1]{\orgname{IIAU-Lab, Dalian University of Technology}, \orgaddress{\city{Dalian}, \country{China}}}
\affil[2]{\orgname{Zhongshan Hospital of Dalian University}, \orgaddress{\city{Dalian}, \country{China}}}
\affil[3]{\orgname{Yale University}, \orgaddress{\country{USA}}}
\affil[4]{\orgname{Nanyang Technological University}, \orgaddress{\country{Singapore}}}


\abstract{Accurate medical image segmentation is critical for early medical diagnosis. Most existing methods are based on U-shape structure and 
use element-wise addition or concatenation to fuse different level features progressively in decoder. 
However, both the two operations easily generate plenty of redundant information, which will weaken the complementarity between different level features, resulting in inaccurate localization and blurred edges of lesions. 
To address this challenge, we propose a general multi-scale in multi-scale subtraction network (M$^{2}$SNet) to finish diverse segmentation from medical image. 
Specifically, we first design a basic subtraction unit (SU) to produce the difference features between adjacent levels in encoder. Next, we expand the single-scale SU to the intra-layer multi-scale SU, which can provide the decoder with both pixel-level and structure-level difference information. Then, we pyramidally equip the multi-scale SUs at different levels with varying receptive fields, thereby achieving the inter-layer multi-scale feature aggregation and obtaining rich multi-scale difference information. In addition, we build a training-free network ``LossNet'' to comprehensively supervise the task-aware features from bottom layer to top layer, which drives our multi-scale subtraction network to capture the detailed and structural cues simultaneously. 
Without bells and whistles, our method performs favorably against most state-of-the-art methods under different evaluation metrics on eleven datasets of four different medical image segmentation tasks of diverse image modalities, including color colonoscopy imaging, ultrasound imaging, computed tomography (CT), and optical coherence tomography (OCT). The source code can be available at \url{https://github.com/Xiaoqi-Zhao-DLUT/MSNet}.
}




\maketitle

\begin{figure*}[t]
\centering
\includegraphics[width=\linewidth]{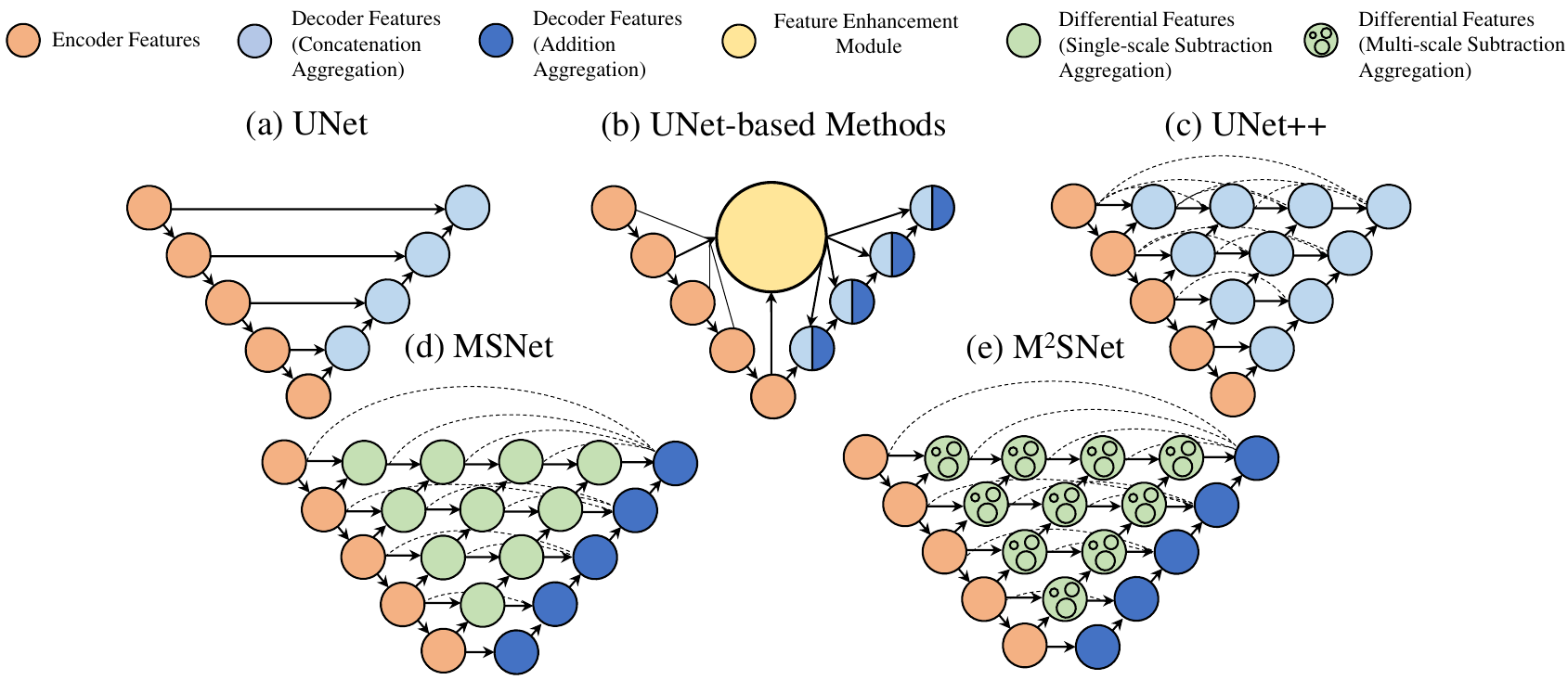}
\caption{Illustration of different medical image segmentation architectures. }
\label{fig:msnet++_msnet_unet_unet++_comparison}
\end{figure*}
\section{Introduction}
\label{sec:introduction}
As 
the important role in computer-aided diagnosis system, accurate medical image segmentation technique can provide the doctors with great guidance for making clinical decisions. There are three general challenges in accurate segmentation: \textbf{Firstly}, U-shape structures~\cite{FPN,UNet} have received considerable attention due to their abilities of utilizing multi-level information to reconstruct high-resolution feature maps. In UNet~\cite{UNet}, the up-sampled feature maps are concatenated with feature maps skipped from the encoder and convolutions and non-linearities are added between up-sampling steps, as shown in Fig.~\ref{fig:msnet++_msnet_unet_unet++_comparison} (a). Subsequent UNet-based methods design diverse feature enhancement modules via attention mechanism~\cite{ResUnet++,PraNet,UniMRSeg}, gate mechanism~\cite{BMPM,GateNet,GateNetv2}, transformer technique~\cite{UTNet,nnWNet,TransUNet_MIA}, as shown in Fig.~\ref{fig:msnet++_msnet_unet_unet++_comparison} (b). UNet++~\cite{UNet++} uses nested and dense skip connections to reduce the  semantic  gap  between  the feature  maps  of  encoder  and  decoder, as shown in Fig.~\ref{fig:msnet++_msnet_unet_unet++_comparison} (c). Generally speaking, different level features in encoder have different characteristics. High-level ones have more semantic information which helps localize the objects, while low-level ones have more detailed information which can capture the subtle boundaries of objects. The  decoder leverages the level-specific and cross-level characteristics to generate the final high-resolution prediction. Nevertheless, the aforementioned methods directly use an element-wise addition or concatenation to fuse any two level features from the encoder and transmit them to the decoder. 
These simple operations do not pay more attention to differential information between different levels. This drawback not only generates redundant information to dilute the really useful features but also weakens the characteristics of level-specific features, which results in that the network can not balance 
accurate localization and subtle boundary refinement.
\textbf{Secondly}, due to the limited receptive field, a single-scale convolutional kernel is difficult to capture context information of size-varying objects. Some methods~\cite{FPN,UNet,UNet++,U2Net,MINet,ZoomNeXt} rely on the inter-layer multi-scale features and progressively integrate the semantic context and texture details from diverse scale representations. Others~\cite{GateNetv2,Spider,UniMRSeg} focus on extracting the intra-layer multi-scale information based on the atrous spatial pyramid pooling module~\cite{ASPP} (ASPP) or DenseASPP~\cite{DenseASPP} in their networks. However, the ASPP-like multi-scale convolution modules will produce many extra parameters and computations. Many methods~\cite{BMPM,UCNet_RGBDSOD,PFNet_COD,BDRAR_Shadow,AFFPN_Shadow} usually equip several ASPP modules into the encoder/decoder blocks of different levels, while some ones~\cite{R3Net,DMRA_RGBDSOD,CoNet_RGBDSOD,Rank-Net_COD} install it on the highest-level encoder block. 
\textbf{Thirdly}, the form of the loss function directly provides the direction for the gradient optimization of the network. In segmentation field, there are many loss functions are proposed to supervise the prediction at the different levels, such as the L1 loss, cross-entropy loss and weighted cross-entropy loss~\cite{FCN} in the pixel level, the SSIM~\cite{SSIM} loss and uncertainty-aware loss~\cite{ZoomNet} in the region level, the IoU loss, Dice loss and consistency-enhanced loss~\cite{MINet} in the  global level. Although these basic loss functions and their variants have different optimization characteristics, the designs of complex manual math forms are really time-consuming for many researches. In order to obtain comprehensive performance, models usually integrate a variety of loss functions, which places great demands on the training skills of the researchers. Therefore, we think that it is necessary to introduce an intelligent loss function without complex manual designs to comprehensively supervise the segmentation prediction.

In this paper, we propose a novel multi-scale in multi-scale subtraction network (M$^{2}$SNet) for general medical image segmentation. 
Firstly, we design a subtraction unit (SU) and apply it to each pair of adjacent level features. The SU highlights the useful difference information between the features and eliminates the interference from the redundant parts.
Secondly, we collect the extreme multi-scale information with the help of the proposed multi-scale in multi-scale subtraction module. For the inter-layer multi-scale information, we pyramidally concatenate multiple subtraction units to capture the large-span cross-level information. Then, we aggregate level-specific features and multi-path cross-level differential features and then generate the final prediction in decoder. For the intra-layer multi-scale information, we improve the single-scale subtraction unit to the multi-scale subtraction unit through a group of full one filters with different kernel sizes, which can achieve naturally multi-scale subtraction aggregation without introducing extra parameters. As shown in Fig.~\ref{fig:msnet++_msnet_unet_unet++_comparison}, MSNet equips the inter-layer multi-scale subtraction module and M$^{2}$SNet has both the inter-layer and intra-layer multi-scale subtraction structures.  
Thirdly, we propose a LossNet to automatically supervise the extracted feature maps from bottom layer to top layer, which can optimize the segmentation from detail to structure with a simple L2-loss function.

Our main contributions are summarized as follows:
\begin{itemize} 
\item  We present a new segmentation framework by replacing traditional addition or concatenation feature fusion with an efficient subtraction aggregation.  
\item  We propose a simple yet general multi-scale in multi-scale subtraction network (M$^{2}$SNet) for diverse medical image segmentation. With multi-scale in multi-scale module, the multi-scale complementary information from lower order to higher order among different levels can be effectively obtained, thereby comprehensively enhancing the perception of organs or lesion areas.
\item  We design an efficient intra-layer multi-scale subtraction unit (MSU). Due to the low parameters and computation of MSU, it can be equipped for all cross-layer aggregations in our M$^{2}$SNet.
\item  We build a general training-free loss network to implement the detail-to-structure supervision in the feature levels, which provides the important supplement to the loss design based on the prediction itself.
\item  We verify the effectiveness of the M$^{2}$SNet on four challenge medical segmentation tasks: polyp segmentation, breast cancer segmentation, lung infection and OCT layer segmentation corresponding to the color colonoscopy imaging, ultrasound imaging, computed tomography (CT), and optical coherence tomography (OCT) image input modality, respectively. In addition, M$^{2}$SNet won the second place in the MICCAI2022 GOALS International Ophthalmology Challenge.
\end{itemize}

\section{Related Work}
\label{sec:relatedwork}
\subsection{{Medical Image Segmentation Network}}\label{subsec:misn}
{Existing medical image segmentation methods can be broadly categorized into medical-general and medical-specific approaches.}
\\
{\textbf{Medicine-general Methods.}
Since the success of U-Net~\cite{UNet}, encoder–decoder architectures with skip connections have become the standard paradigm. Variants such as U-Net++~\cite{UNet++} and Attention U-Net~\cite{Attention_UNet} improve feature aggregation and target localization through redesigned skip pathways or attention mechanisms.
More recently, Transformer-based models~\cite{SegFormer,SwinUNet,UTNet,TransUNet_MIA} introduce self-attention to capture long-range dependencies, enhancing global context modeling. However, their quadratic complexity and relatively weak local inductive bias may limit scalability and fine-grained prediction.
To address efficiency issues, state-space–based architectures such as Mamba~\cite{Mamba} and its segmentation variants~\cite{Sigma,Segmamba,U-mamba} replace attention with structured recurrence, achieving linear complexity while maintaining long-range modeling. Hybrid CNN–Mamba frameworks further balance global context modeling and local detail preservation. 
}
\\
{\textbf{Medicine-specific Methods.} 
Task-oriented designs tailor architectures to specific organs or lesions. For example, polyp segmentation models~\cite{SFA,PraNet,UM-Net,PNSNet} emphasize boundary refinement and contextual modeling. Lung infection segmentation methods~\cite{Anam-Net,Inf-Net,BCS-Net} focus on edge modeling and anomaly localization, while breast tumor segmentation approaches~\cite{SKUNet,NU-net} enhance receptive field adaptation and representation robustness.
Overall, medical-general methods target universal challenges such as multi-scale representation and cross-level feature aggregation, whereas medical-specific models introduce task-dependent modules, including attention mechanisms and boundary enhancement strategies. Nevertheless, most existing approaches rely heavily on addition or concatenation for feature fusion, which may dilute discriminative difference information. In contrast, our multi-scale subtraction module explicitly models feature discrepancies, enabling more targeted and efficient representation learning.}

{We can see that the medicine-general methods are usually towards general challenges (i.e., rich feature representation, multi-scale information extraction and cross-level feature aggregation). And, the medicine-specific methods propose targeted solutions based on the characteristics of the current organ or lesion, such as designing a series of attention mechanisms, edge enhancement modules, uncertainty estimation, etc. 
However, both general medicine-general and medicine-specific models rely on a large number of addition or concatenation operations to achieve feature fusion, which weakens the specificity parts among complementary features. Our proposed multi-scale subtraction module naturally focuses on extracting difference information, thus providing the decoder with efficient targeted features.}

\subsection{{Multi-scale Feature Extraction}}\label{subsec:multiscale}
{Multi-scale modeling is essential for handling objects with large scale variations in medical images. Existing strategies can be grouped into inter-layer and intra-layer designs. 
\textbf{Inter-layer multi-scale structures} aggregate features from different encoder stages within U-shaped frameworks~\cite{UNet,UNet++,FPN,PraNet,U2Net}. These architectures progressively fuse hierarchical features to combine semantic and spatial information.
\textbf{Intra-layer multi-scale structures} employ parallel convolutional branches with different receptive fields, such as ASPP~\cite{ASPP}, DenseASPP~\cite{DenseASPP}, and related variants, to capture contextual information at multiple scales.
Different from prior work that focuses primarily on feature enrichment via aggregation, we incorporate subtraction operations into both inter-layer and intra-layer multi-scale fusion. By explicitly extracting difference information rather than merely accumulating features, our design enhances discriminative representation while maintaining structural simplicity.}
\begin{figure*}[t]
\centering
\includegraphics[width=\linewidth]{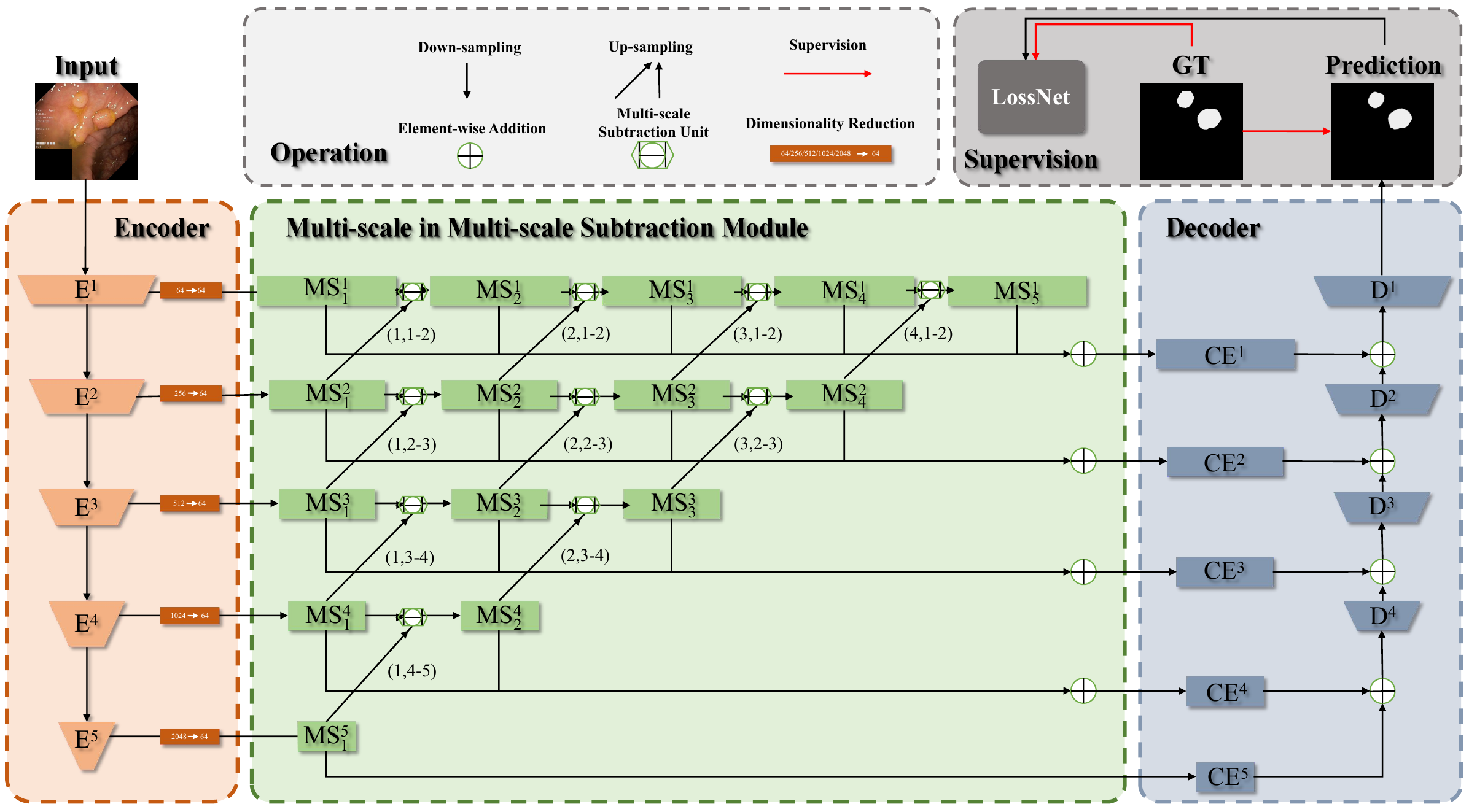}
\caption{Overview of the proposed multi-scale subtraction network.}
\label{fig:Pipeline}
\end{figure*}
\subsection{{Segmentation Loss Functions and Perceptual Supervision}}\label{subsec:loss}
Most loss functions in image segmentation are based on cross-entropy or coincidence measures. The traditional cross-entropy loss treats the categories information equally.  Long \textit{et al.}~\cite{FCN} propose a weighted cross-entropy loss (WCE) for each class to offset the class imbalance in the data. Lin \textit{et al.}~\cite{Focal-loss} introduce the weights of difficult and easy samples to propose the Focal loss.  Dice loss~\cite{V-net} is proposed as the loss function of coincidence measurement in V-Net, which can effectively suppress problems caused by category imbalance. Tversky loss~\cite{Tversky-loss} is a regularized version of Dice loss to control the contribution of accuracy and recall to the loss function. Wong \textit{et al.}~\cite{EL-LOSS} propose exponential logarithmic loss (EL Loss) through the weighted summation of Dice loss and WCE loss to improve the segmentation accuracy of small structure objects. Taghanaki \textit{et al.}~\cite{Combo-loss} find that there is a risk in using the loss function based on overlap alone, and propose the como-loss to combine Dice loss as a regularization term with WCE loss to deal with the problem of input and output imbalance. Although these various loss functions have different effects at different levels, it is indeed time-consuming and laborious to manually design these complex functions. To this end, we propose the automatic and comprehensive segmentation loss structure, coined as the LossNet.

\section{Method}
The M$^{2}$SNet architecture is shown in Fig.~\ref{fig:Pipeline}, in which there are five encoder blocks ($\mathbf{E}^i$, $i \in \left \{1, 2, 3, 4, 5 \right \}$), a multi-scale in multi-scale subtraction module (MMSM) and four decoder blocks ($\mathbf{D}^i$, $i \in \left \{1, 2, 3, 4 \right \}$). We adopt the Res2Net-50 as the backbone to extract five levels of features. First, we separately adopt a $3 \times 3$ convolution for feature maps of each encoder block to reduce the channel to $64$, which can decrease the number of parameters for subsequent operations. Next, these different level features are fed into the MMSM and output five complementarity 
enhanced features (${CE}^i$, $i \in \left \{1, 2, 3, 4, 5 \right \}$). Finally, each ${CE}^i$ progressively participates in the decoder and generates the final prediction. In the training phase, both the prediction and ground truth are input into the LossNet to achieve supervision. We describe the multi-scale in multi-scale subtraction module in Sec.~\ref{sec:msm} and give the details of LossNet in Sec.~\ref{sec:lossnet}.
\subsection{Multi-scale in Multi-scale Subtraction Module}\label{sec:msm}
\begin{figure}[t]
\centering
\includegraphics[width=1.0\linewidth]{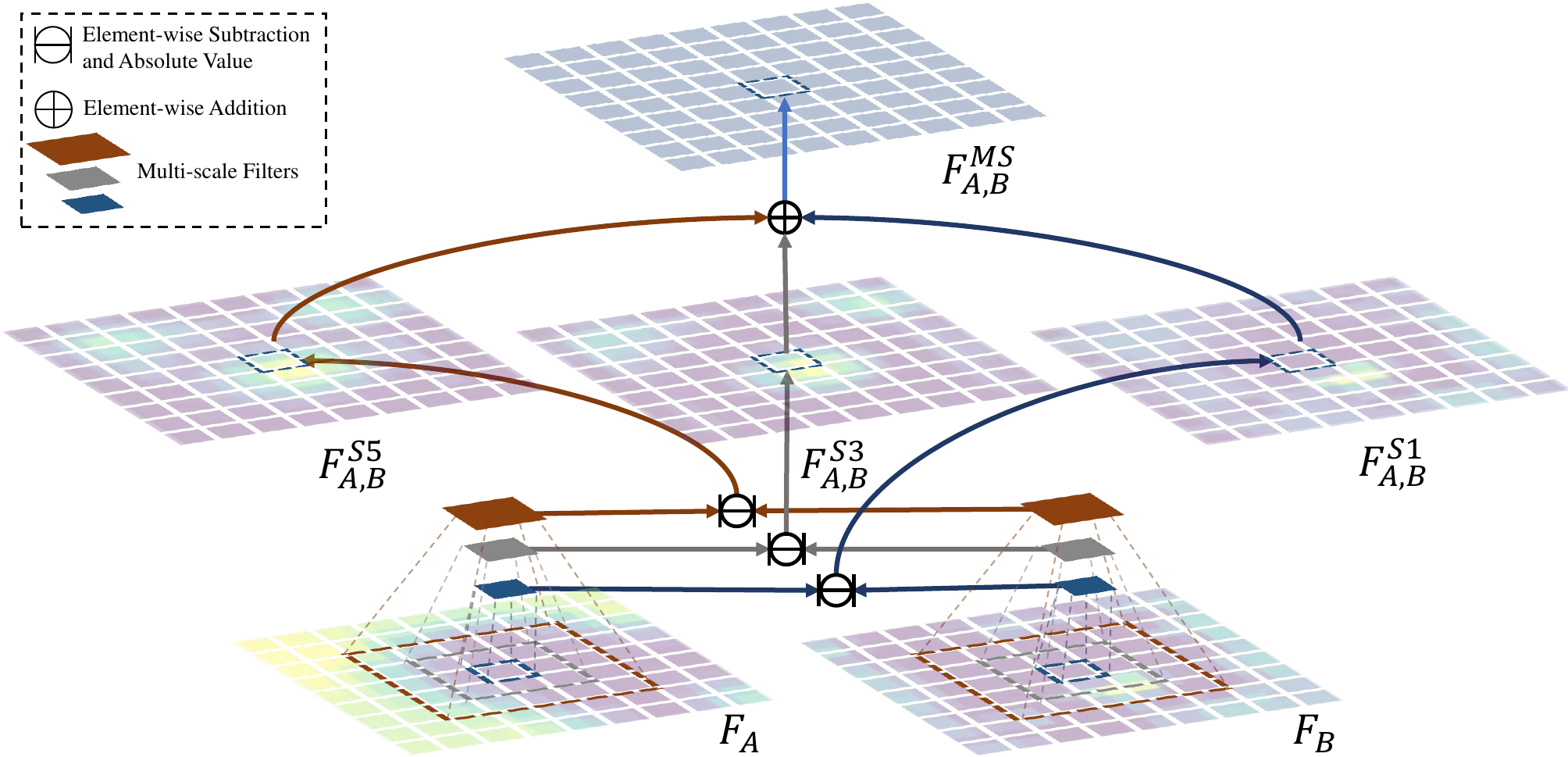}
\caption{Detailed diagram of multi-scale subtraction unit.}
\label{fig:MSU}
\end{figure}

We use  $F_{A}$ and $F_{B}$ to represent adjacent level feature maps. They all have been activated by the ReLU operation. We  define a basic subtraction unit (SU):
  \begin{equation}\label{equ:1}
 \begin{split}
     SU = Conv(\vert F_{A} \ominus F_{B} \vert ),
 \end{split}
\end{equation}
 where $\ominus$ is the element-wise subtraction operation, $ \vert \cdot \vert$ calculates the absolute value and $Conv(\cdot)$ denotes the convolution layer. Directly performing single-scale subtraction on the features of element positions is only to establish the difference relationship on the isolated pixel level, without considering that the lesion may have the characteristics of regional clustering. Compared to the MICCAI version~\cite{MSNet} of MSNet with the single-scale subtraction unit, we design a powerful intra-layer multi-scale subtraction unit (MSU) and improve MSNet to M$^{2}$SNet. As shown in Fig.~\ref{fig:MSU}, we utilize the multi-scale convolution filters with fixed full one weights of size $1 \times 1$, $3 \times 3$ and $5 \times 5$ to calculate the detail and structure difference values according to the pixel-pixel and region-region pattern. Using multi-scale filters with fixed parameters not only can directly capture the multi-scale difference clues between initial feature pairs at matched spatial locations, but also achieve efficient training without introducing additional parameter burdens. Therefore, M$^{2}$SNet can maintain the same low computation as MSNet and achieve higher precision performance. The entire multi-scale subtraction process can be formulated as: 
\begin{equation}\label{equ:2}
  \begin{split}
     MSU = Conv(  \quad \quad \quad \quad \quad \quad     \\
     \vert Filter( F_{A} )_{1\times1}\ominus Filter( F_{B} )_{1\times1} \vert + \\ \vert Filter( F_{A} )_{3\times3}\ominus Filter( F_{B} )_{3\times3} \vert + \\ \vert Filter( F_{A} )_{5\times5}\ominus Filter( F_{B} )_{5\times5} \vert ),
 \end{split}
 \end{equation}
 where $Filter(\cdot)_{n \times n}$ represents the full one filter of size $n \times n$. The MSU can capture the complementary information of $F_{A}$ and $F_{B}$ and highlight their differences from texture to structure, thereby providing richer information for the decoder.

To obtain higher-order complementary information across multiple feature levels,  
we horizontally and vertically concatenate multiple MSUs to calculate a series of differential features with different orders and receptive fields. The detail of the multi-scale in multi-scale subtraction module can be found in Fig.~\ref{fig:Pipeline}. We aggregate the scale-specific feature ($MS^{i}_{1}$) and cross-scale differential features ($MS^{i}_{n \neq 1}$) between the corresponding level and any other levels to generate complementarity enhanced feature ($CE^{i}$). This process can be formulated as follows:
\begin{equation}\label{equ:3}
 \begin{split}
     CE^{i} = Conv(\sum_{n=1}^{6-i}MS^{i}_{n} ) \quad i=1, 2, 3, 4, 5.
 \end{split}
\end{equation}
Finally, all $CE^{i}$ participate in decoding and then the polyp region is segmented.  
\subsection{LossNet}\label{sec:lossnet}

\begin{figure}[t]
\centering
\includegraphics[width=\linewidth]{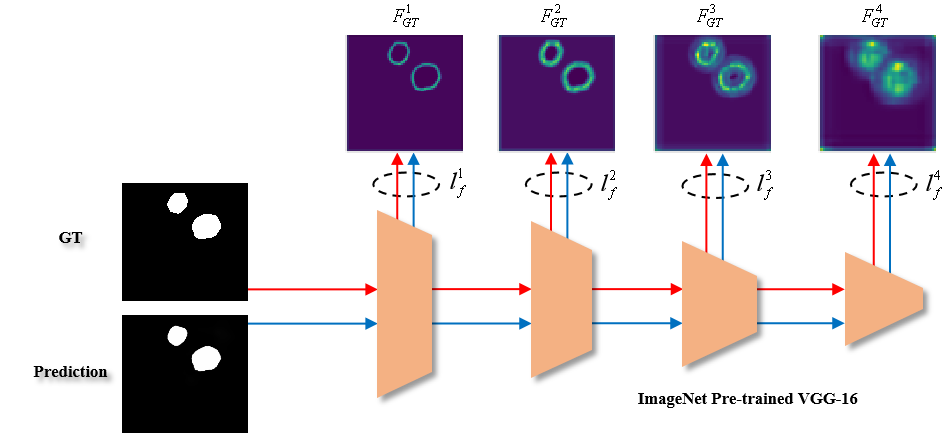}
\caption{Illustration of LossNet.}
\label{fig:P_loss}
\end{figure}

In the proposed model, the total training loss can be written as:
\begin{equation}\label{equ:4}
 \begin{split}
    \mathcal{L}_{total}=\mathcal{L}_{IoU}^w +\mathcal{L}_{BCE}^w + \mathcal{L}_{f},
 \end{split}
\end{equation}
 where $\mathcal{L}_{IoU}^w$ and $\mathcal{L}_{BCE}^w$ represent the weighted IoU loss and binary cross-entropy (BCE) loss which have been widely adopted in segmentation tasks. We use the same definitions as in ~\cite{PraNet,F3Net,BASNet} and their effectiveness has been validated in these works. Different from them, we extra use a LossNet to further optimize the segmentation from detail to structure. Specifically, we use an ImageNet pre-trained classification network, such as VGG-16, to extract the multi-scale features of the prediction and ground truth, respectively. Then, their feature difference is computed as loss $\mathcal{L}_{f}$: 
\begin{equation}\label{equ:5}
 \begin{split}
    \mathcal{L}_{f} = {l}_{f}^{1} + {l}_{f}^{2} + {l}_{f}^{3} + {l}_{f}^{4}.
 \end{split}
\end{equation}
Let ${F}_{P}^{i}$ and ${F}_{G}^{i}$ separately represent the $i$-th level feature maps extracted from the prediction and ground truth. The ${l}_{f}^{i}$ is calculated as their Euclidean distance (L2-Loss), which is supervised at the pixel level:
\begin{equation}\label{equ:5}
 \begin{split}
    {l}_{f}^{i} = \vert\vert {F}_{P}^{i} -  {F}_{G}^{i} \vert\vert_{2}, \quad i=1, 2, 3, 4.
 \end{split}
\end{equation}
As can be seen from Fig~\ref{fig:P_loss}, the low-level feature maps contain rich boundary information and the high-level ones depict location information. Thus, the LossNet can generate comprehensive supervision at the feature levels. 

\section{Experiments}
\subsection{Datasets}
Extensive experiments are conducted to verify the effectiveness of the proposed framework on four different types of medical segmentation tasks with data from varied image modalities, including color colonoscopy imaging, ultrasound imaging, computed tomography (CT), and optical coherence tomography (OCT).

\textbf{Polyp Segmentation.}
According to GLOBOCAN 2020 data, colorectal cancer is the third most common cancer worldwide and the second most common cause of death. It usually begins as small, noncancerous (benign) clumps of cells called polyps that form on the inside of the colon. We evaluate the proposed model on five benchmark datasets: CVC-ColonDB~\cite{CVC-ColonDB}, ETIS~\cite{ETIS}, Kvasir~\cite{Kvasir}, CVC-T~\cite{CVC-T} and CVC-ClinicDB~\cite{CVC-ClinicDB}. We adopt the same training set as the latest image polyp segmentation method~\cite{PraNet}, that is, $900$ samples from the Kvasir and $550$ samples from the CVC-ClinicDB~\cite{CVC-ClinicDB} are used for training. 
The remaining images and the other three datasets are used for testing. Besides, there are some video-based polyp datasets, including the CVC-300\cite{cvc-300} and CVC-612\cite{CVC-ClinicDB}. We follow the latest video polyp segmentation method to split the videos from CVC-300 (12 clips) and CVC-612 (29 clips)
into 60\% for training, 20\% for validation, and 20\% for testing.

\textbf{Lung Infection.}
A novel viral pneumonia that emerged in early 2020 rapidly spread worldwide, creating an unprecedented public-health challenge. At present, only a few public lung CT datasets are available for infection-area segmentation. To obtain a relatively sufficient sample size for training, we slice one publicly available dataset~\cite{COVID-19_dataset1} and merge it with another public dataset~\cite{COVID-19_dataset2}, resulting in 1,277 high-quality CT images through uniform sampling. These are then split into 894 images for training and 383 images for testing.

\textbf{Breast Ultrasound Segmentation.}
Breast cancer is one of the most dreaded cancers in 
women~\cite{Breast_cancer}. Segmenting the lesion region from 
breast ultrasound images is essential for tumor diagnosis. BUSI~\cite{BUSI} dataset contains $780$ images of $600$ female patients. Among them, there are $133$ normal cases, $437$ benign tumors, and $210$ malignant tumors. We follow the popular breast ultrasound segmentation methods~\cite{SKUNet,NU-net} to perform four-fold cross-validation on BUSI. 

\begin{figure}[t]
\centering
\includegraphics[width=\linewidth]{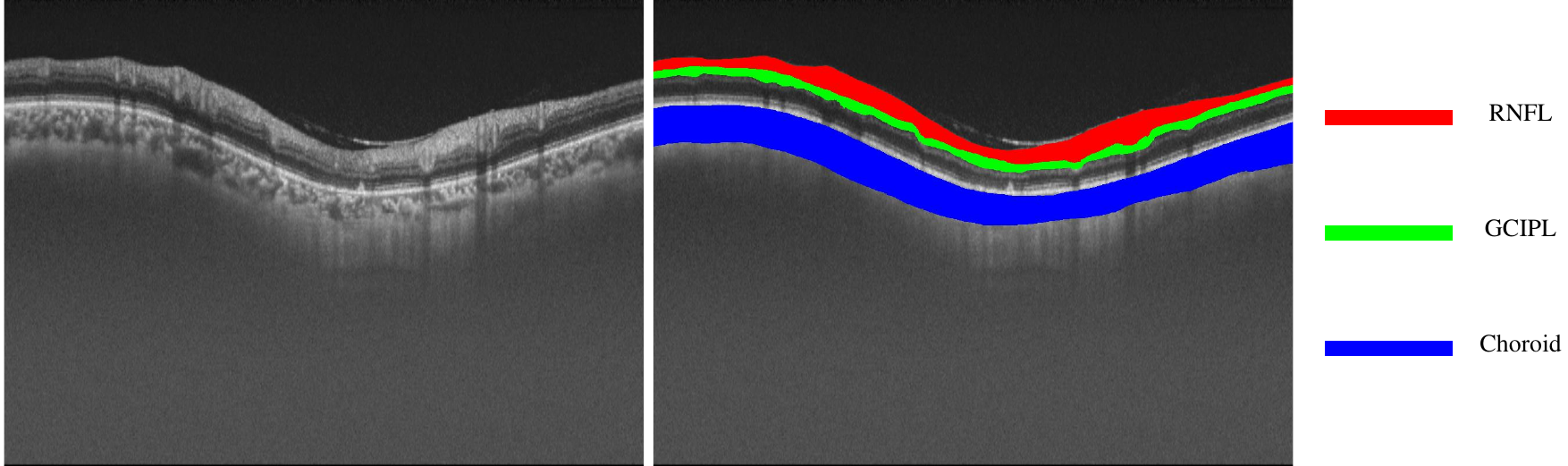}
\caption{Visualization of the fundus OCT layer segmentation.}
\label{fig:GOALS_show}
\end{figure}

\textbf{OCT Layer Segmentation.} The OCT images are often used to diagnose and monitor retinal diseases more accurately based on abnormality quantification and retinal layer thickness computation both in research centers and clinic routines. At present, many scholars have been studying the segmentation of fundus structure in macular OCT scans, but few focus on parapapillary circular scans. To fully show the generalization of M$^{2}$SNet in different medical tasks, we take our M$^{2}$SNet to participate in the \textcolor{orange}{MICCAI 2022 Challenge: Glaucoma Oct Analysis and Layer Segmentation (GOALS)}\footnote{\url{https://conferences.miccai.org/2022/en/MICCAI2022-CHALLENGES.html}}. 
It requests participants to segment three layers, which has positive significance for the diagnosis of glaucoma, including retinal nerve fiber layer (RNFL), ganglion cell-inner plexiform layer (GCIPL), and choroid layer, as shown in Fig.~\ref{fig:GOALS_show}. The GOALS2022~\cite{GOALS} dataset~\cite{GOALS_OMIA} contains $300$ circumpapillary OCT. There are three equal groups with $100$ OCT images for the training process, the preliminary competition process and the final process, respectively. The GOALS2022 challenge attracts $100$ teams from all over the world to participate, and we finally won \textcolor{orange}{the second place (2/100)}\footnote{\url{https://github.com/Xiaoqi-Zhao-DLUT/MSNet}}.

\subsection{Evaluation Metrics}
There are many popular metrics used in different medical segmentation branches. mean Dice (mDice), mean IoU (mIoU), the weighted F-measure ($F_{\beta}^{w}$)~\cite{Fwb}, S-measure ($S_{\alpha}$)~\cite{S-m}, E-measure ($E_\phi^{max}$)~\cite{Em} and mean absolute error (MAE) are widely used in polyp segmentation. Following~\cite{Inf-Net}, five metrics are employed for quantitative evaluation, including Precision, Recall, Dice
Similarity Coefficient (DSC)~\cite{DSC}, S-measure and MAE. Jaccard, Precision, Recall, Dice and Specificity~\cite{AAU-net,NU-net} are more commonly used for breast tumor segmentation. For OCT layer segmentation, GOALS2022~\cite{GOALS_OMIA} adopts the Dice coefficient and mean Euclidean distance (MED) to evaluate segmentation bodies and edges, respectively.  The lower value is better for the MAE and MED, and higher is better for others.

\begin{table}[t]
  \centering
  \scriptsize
  \renewcommand{\arraystretch}{1.1}
  \setlength\tabcolsep{5pt}
  \caption{{Quantitative comparisons on image polyp segmentation datasets. Top $2$ scores are highlighted in \textBC{mycellreda}{red} and \textBC{mycellblue}{blue}, respectively.  ``$\dagger$'' represents the medicine-specific method. }
  }\label{tab:image_polyp_comparison}
  \resizebox{\columnwidth}{!}
  {
 \begin{tabular}{cr|c||ccccccc}
  \hline
  \rowcolor{mygray}
  &Methods & Backbone&mDice $\uparrow$ & mIoU $\uparrow$  &  $F_\beta^w$  $\uparrow$& $S_{\alpha}$ $\uparrow$&$E_\phi^{max}$ $\uparrow$ & MAE $\downarrow$\\
  \hline
  \multirow{9}{*}{\begin{sideways}ColonDB\end{sideways}} & 
 U-Net~\cite{UNet}  & R2-50&0.519 & 0.449 & 0.498 & 0.711 & 0.763 & 0.061  \\
&U-Net++~\cite{UNet++} & R2-50& 0.490 & 0.413 & 0.467 & 0.691 & 0.762 & 0.064 \\
&Atten-UNet~\cite{Attention_UNet} & R2-50& 0.466 & 0.385 & 0.431 & 0.670 & 0.724 & 0.071 \\
&UTNet~\cite{UTNet} & R-50 + ViT-B16& 0.676 & 0.600 & 0.656 & 0.799 & 0.855 & 0.041  \\
&TransUnet~\cite{TransUNet} & R-50 + ViT-B16& 0.717 & 0.645 & 0.685 & 0.824 & 0.841 & 0.044  \\
&SFA$^{\dagger}$~\cite{SFA}& R2-50 & 0.467 & 0.351 & 0.379 & 0.634 & 0.648 & 0.094 \\
&PraNet$^{\dagger}$~\cite{PraNet} & R2-50& 0.716 & 0.645 & \cellcolor{mycellblue}\textBC{white}{0.699} & 0.820 & 0.847 & 0.043 \\
\hline
&MSNet~\cite{MSNet} & R2-50& \cellcolor{mycellblue}\textBC{white}{0.755} & \cellcolor{mycellblue}\textBC{white}{0.678} & \cellcolor{mycellreda}\textBC{white}{0.737} & \cellcolor{mycellblue}\textBC{white}{0.836} & \cellcolor{mycellreda}\textBC{white}{0.883} & \cellcolor{mycellblue}\textBC{white}{0.041}\\
&{M$^{2}$SNet}  & R2-50& \cellcolor{reda}\textBC{white}{0.758} & \cellcolor{reda}\textBC{white}{0.685} & \cellcolor{mycellreda}\textBC{white}{0.737} & \cellcolor{mycellreda}\textBC{white}{0.842} &\cellcolor{mycellblue}\textBC{white}{0.869} & \cellcolor{mycellreda}\textBC{white}{0.038}\\
  \hline
  \hline
  \multirow{9}{*}{\begin{sideways}ETIS\end{sideways}} &
 U-Net~\cite{UNet}  & R2-50& 0.406 & 0.343 & 0.366 & 0.682 & 0.645 & 0.036\\
&U-Net++~\cite{UNet++} & R2-50 & 0.413 & 0.342 & 0.390 & 0.681 & 0.704 & 0.035 \\
&Atten-UNet~\cite{Attention_UNet} & R2-50& 0.382 & 0.308 & 0.372 & 0.641 & 0.670 & 0.050\\
&UTNet~\cite{UTNet}  & R-50 + ViT-B16& 0.556 & 0.489 & 0.522 & 0.749 & 0.772 & 0.022 \\
&TransUNet~\cite{TransUNet} & R-50 + ViT-B16 & 0.573 & 0.512 & 0.517 & 0.765 & 0.707 & 0.029 \\
&SFA$^{\dagger}$~\cite{SFA}  & R2-50& 0.297 & 0.219 & 0.231 & 0.557 & 0.515 & 0.109 \\
&PraNet$^{\dagger}$~\cite{PraNet}  & R2-50& 0.630 & 0.576 & 0.600 & 0.791 & 0.792 & 0.031\\
\hline
&MSNet~\cite{MSNet}  & R2-50& \cellcolor{mycellblue}\textBC{white}{0.719} &\cellcolor{mycellblue}\textBC{white}{0.664} & \cellcolor{mycellblue}\textBC{white}{0.678} & \cellcolor{mycellblue}\textBC{white}{0.840} & \cellcolor{mycellblue}\textBC{white}{0.830} & \cellcolor{mycellblue}\textBC{white}{0.020}\\
&{M$^{2}$SNet}   & R2-50& \cellcolor{mycellreda}\textBC{white}{0.749} & \cellcolor{mycellreda}\textBC{white}{0.678} & \cellcolor{mycellreda}\textBC{white}{0.712} & \cellcolor{mycellreda}\textBC{white}{0.846}  & \cellcolor{mycellreda}\textBC{white}{0.872} & \cellcolor{mycellreda}\textBC{white}{0.017}\\
  \hline
  \hline
  \multirow{9}{*}{\begin{sideways}Kvasir\end{sideways}} &
U-Net~\cite{UNet}   & R2-50& 0.821 & 0.756 & 0.794 & 0.858 & 0.901 & 0.055 \\
&U-Net++~\cite{UNet++}  & R2-50& 0.824 & 0.753 &0.808& 0.862 & 0.907 &  0.048 \\
&Atten-UNet~\cite{Attention_UNet} & R2-50 & 0.769 & 0.683 & 0.730 & 0.828 & 0.859 & 0.062 \\
&UTNet~\cite{UTNet} & R-50 + ViT-B16& 0.862 & 0.803 & 0.843 & 0.886 & 0.911 & 0.042 \\
&TransUNet~\cite{TransUNet}   & R-50 + ViT-B16& 0.869 & 0.816 & 0.847 & 0.899 & 0.920 & 0.040 \\
&SFA$^{\dagger}$~\cite{SFA}& R2-50& 0.725 & 0.619 & 0.670 & 0.782 & 0.828 & 0.075\\
&PraNet$^{\dagger}$~\cite{PraNet} & R2-50 & 0.901 & 0.848 & 0.885 & \textBC{blue}{0.915} & 0.943 & 0.030\\
\hline
&MSNet~\cite{MSNet} & R2-50& \cellcolor{mycellblue}\textBC{white}{0.907} & \cellcolor{mycellreda}\textBC{white}{0.862} & \cellcolor{mycellblue}\textBC{white}{0.893} & \cellcolor{mycellreda}\textBC{white}{0.922} & \cellcolor{mycellblue}\textBC{white}{0.944} & \cellcolor{mycellblue}\textBC{white}{0.028}\\
&{M$^{2}$SNet} & R2-50& \cellcolor{mycellreda}\textBC{white}{0.912} & \cellcolor{mycellblue}\textBC{white}{0.861} & \cellcolor{mycellreda}\textBC{white}{0.901} & \cellcolor{mycellreda}\textBC{white}{0.922} & \cellcolor{mycellreda}\textBC{white}{0.953} & \cellcolor{mycellreda}\textBC{white}{0.025}\\
  \hline
  \hline
  \multirow{9}{*}{\begin{sideways}CVC-T\end{sideways}} &
U-Net~\cite{UNet} & R2-50 & 0.717 & 0.639 & 0.684 & 0.842 & 0.867 & 0.022\\
&U-Net++~\cite{UNet++} & R2-50 & 0.714 & 0.636 &0.687& 0.838 & 0.884 &  0.018 \\
&Atten-UNet~\cite{Attention_UNet} & R2-50 & 0.603 & 0.511 & 0.554 & 0.760 & 0.819 & 0.024 \\
&UTNet~\cite{UTNet}  & R-50 + ViT-B16& 0.806 & 0.733 & 0.778 & 0.882 & 0.924 & 0.016 \\
&TransUNet~\cite{TransUNet}  & R-50 + ViT-B16& 0.828 & 0.757 & 0.785 & 0.906 & 0.901 & 0.015 \\
&SFA$^{\dagger}$~\cite{SFA}  & R2-50& 0.465 & 0.332 & 0.341 & 0.640 & 0.604 & 0.065\\
&PraNet$^{\dagger}$~\cite{PraNet} & R2-50& \cellcolor{mycellblue}\textBC{white}{0.873} & 0.804 & 0.843 & 0.924 & 0.938 &\cellcolor{mycellblue}\textBC{white}{0.010}\\
\hline
&MSNet~\cite{MSNet}& R2-50 & 0.869 & \cellcolor{mycellblue}\textBC{white}{0.807} & \cellcolor{mycellblue}\textBC{white}{0.849} & \cellcolor{mycellblue}\textBC{white}{0.925} & \cellcolor{mycellblue}\textBC{white}{0.943} & \cellcolor{mycellblue}\textBC{white}{0.010}\\
&{M$^{2}$SNet} & R2-50 & \cellcolor{mycellreda}\textBC{white}{0.903} & \cellcolor{mycellreda}\textBC{white}{0.842} & \cellcolor{mycellreda}\textBC{white}{0.881} & \cellcolor{mycellreda}\textBC{white}{0.939} & \cellcolor{mycellreda}\textBC{white}{0.965} & \cellcolor{mycellreda}\textBC{white}{0.009}\\
  \hline
  \hline
  \multirow{9}{*}{\begin{sideways}ClinicDB\end{sideways}} &
U-Net~\cite{UNet} & R2-50 & 0.824 & 0.767 & 0.811 & 0.889 & 0.917 & 0.019\\
&U-Net++~\cite{UNet++}  & R2-50 & 0.797 & 0.741 &0.785& 0.872 & 0.898 &  0.022\\
&Atten-UNet~\cite{Attention_UNet}& R2-50 & 0.866 & 0.809 & 0.856 & 0.908 & 0.960 & 0.015 \\
&UTNet~\cite{UTNet}  & R-50 + ViT-B16& 0.860 & 0.818 & 0.856 & 0.910 & 0.963 & 0.017 \\
&TransUNet~\cite{TransUNet}  & R-50 + ViT-B16& 0.847 & 0.798 & 0.831 & 0.907 & 0.920 & 0.020 \\
&SFA$^{\dagger}$~\cite{SFA} & R2-50 & 0.698 & 0.615 & 0.647 & 0.793 & 0.816 & 0.042\\
&PraNet$^{\dagger}$~\cite{PraNet} & R2-50 & 0.902 & 0.858 & 0.896 & 0.935 & 0.958 & \cellcolor{mycellblue}\textBC{white}{0.009}\\
\hline
&MSNet~\cite{MSNet} & R2-50 & \cellcolor{mycellblue}\textBC{white}{0.921} & \cellcolor{mycellblue}\textBC{white}{0.879} & \cellcolor{mycellblue}\textBC{white}{0.914} & \cellcolor{mycellblue}\textBC{white}{0.941} & \cellcolor{mycellreda}\textBC{white}{0.972} & \cellcolor{mycellreda}\textBC{white}{0.008}\\
&{M$^{2}$SNet}  & R2-50 & \cellcolor{mycellreda}\textBC{white}{0.922} &\cellcolor{mycellreda}\textBC{white}{0.880} & \cellcolor{mycellreda}\textBC{white}{0.917} & \cellcolor{mycellreda}\textBC{white}{0.942} & \cellcolor{mycellblue}\textBC{white}{0.970} & \cellcolor{mycellblue}\textBC{white}{0.009} \\
  \hline
  \end{tabular}
  }
\end{table}

\begin{table}[!t]
\centering
\scriptsize
  \renewcommand{\arraystretch}{1.1}
\setlength\tabcolsep{5pt}
\caption{ Quantitative comparisons on video polyp segmentation datasets. Top $2$ scores are highlighted in \textBC{mycellreda}{red} and \textBC{mycellblue}{blue}, respectively.  ``$\dagger$'' and ``$\bigstar$'' represent the medicine-specific method and the video polyp method, respectively. 
  }
  \label{tab:video_polyp_comparison}
\resizebox{\columnwidth}{!}
{
\begin{tabular}{cr|c||ccccc}
\hline
\rowcolor{mygray}
&Methods &Backbone& mDice $\uparrow$ & mIoU $\uparrow$ & $S_{\alpha}$ $\uparrow$&$E_\phi^{max}$ $\uparrow$ & MAE $\downarrow$\\
\hline
\multirow{8}{*}{\begin{sideways}CVC-300-TV\end{sideways}} & 
U-Net~\cite{UNet}& R2-50 & 0.639 & 0.525  & 0.793 & 0.826 & 0.027 \\
&U-Net++~\cite{UNet++} & R2-50& 0.649 & 0.539  & 0.796 & 0.831 & 0.024 \\
&ResUNet++$^{\dagger}$~\cite{ResUnet++} & R2-50& 0.535 & 0.412  & 0.703 & 0.718 & 0.052 \\
&ACSNet$^{\dagger}$~\cite{ACSNet} & R2-50& 0.738 & 0.632  & 0.837 & 0.871 & 0.016\\
&PraNet$^{\dagger}$~\cite{PraNet}& R2-50 & 0.739 & 0.645  & 0.833 & 0.852 & 0.016\\
&PNS-Net$^{\dagger\bigstar}$~\cite{PNSNet} & R2-50& \cellcolor{mycellblue}\textBC{white}{0.863} & \cellcolor{mycellblue}\textBC{white}{0.805}  &\cellcolor{mycellblue}\textBC{white}{0.909} &\cellcolor{mycellblue}\textBC{white}{0.921} & \cellcolor{mycellblue}\textBC{white}{0.013}\\
\hline
&MSNet~\cite{MSNet}& R2-50 & 0.837 & 0.755  & 0.895 & 0.942 & 0.014\\
&{M$^{2}$SNet} & R2-50& \cellcolor{mycellreda}\textBC{white}{0.876} &\cellcolor{mycellreda}\textBC{white}{0.805} & \cellcolor{mycellreda}\textBC{white}{0.918} & \cellcolor{mycellreda}\textBC{white}{0.963} &\cellcolor{mycellreda}\textBC{white}{0.010}\\
\hline
\hline
\multirow{8}{*}{\begin{sideways}CVC-612-V\end{sideways}} &
U-Net~\cite{UNet} & R2-50 & 0.725 & 0.610 & 0.826 & 0.855 & 0.023\\
&U-Net++~\cite{UNet++} & R2-50  & 0.684 & 0.570 & 0.805 & 0.830 &  0.025\\
&ResUNet++$^{\dagger}$~\cite{ResUnet++}& R2-50 & 0.752 & 0.648  & 0.829 & 0.877 & 0.023 \\
&ACSNet$^{\dagger}$~\cite{ACSNet}& R2-50 & 0.804 & 0.712  & 0.847 & 0.887 & 0.054\\
&PraNet$^{\dagger}$~\cite{PraNet}& R2-50 & 0.869 & 0.799  & 0.915 & 0.936 & 0.013\\
&PNS-Net$^{\dagger\bigstar}$~\cite{PNSNet} & R2-50& 0.859 & 0.804  & 0.923 & 0.944 & 0.012\\
\hline
&MSNet~\cite{MSNet}& R2-50 & \cellcolor{mycellblue}\textBC{white}{0.889} & \cellcolor{mycellblue}\textBC{white}{0.834} & \cellcolor{mycellblue}\textBC{white}{0.931} & \cellcolor{mycellblue}\textBC{white}{0.959} & \cellcolor{mycellreda}\textBC{white}{0.009}\\
&{M$^{2}$SNet} & R2-50 & \cellcolor{mycellreda}\textBC{white}{0.897} &\cellcolor{mycellreda}\textBC{white}{0.838} & \cellcolor{mycellreda}\textBC{white}{0.936} & \cellcolor{mycellreda}\textBC{white}{0.966} &\cellcolor{mycellblue}\textBC{white}{0.010}\\
\hline
\hline
\multirow{8}{*}{\begin{sideways}CVC-612-T\end{sideways}} &
U-Net~\cite{UNet} & R2-50 & 0.729 & 0.635 & 0.810 & 0.836 & 0.058\\
&U-Net++~\cite{UNet++} & R2-50  & 0.740 & 0.635 & 0.800 & 0.817 &  0.059\\
&ResUNet++$^{\dagger}$~\cite{ResUnet++} & R2-50& 0.617 & 0.514 & 0.727 & 0.758 & 0.084 \\
&ACSNet$^{\dagger}$~\cite{ACSNet}& R2-50 & 0.782 & 0.700  & 0.838 & 0.864 & 0.053\\
&PraNet$^{\dagger}$~\cite{PraNet}& R2-50 & \cellcolor{mycellreda}\textBC{white}{0.852} & \cellcolor{mycellblue}\textBC{white}{0.786} & 0.886 & \cellcolor{mycellblue}\textBC{white}{0.904} & \cellcolor{mycellblue}\textBC{white}{0.038}\\
&PNS-Net$^{\dagger\bigstar}$~\cite{PNSNet} & R2-50& 0.841 & \cellcolor{mycellreda}\textBC{white}{0.788}  & \cellcolor{mycellreda}\textBC{white}{0.903} & 0.903 & \cellcolor{mycellblue}\textBC{white}{0.038}\\
\hline
&MSNet~\cite{MSNet} & R2-50& 0.824 & 0.761  & 0.879 & 0.904 & 0.040\\
&{M$^{2}$SNet} & R2-50 & \cellcolor{mycellblue}\textBC{white}{0.846} &0.782 & \cellcolor{mycellblue}\textBC{white}{0.894} & \cellcolor{mycellreda}\textBC{white}{0.921} &\cellcolor{mycellreda}\textBC{white}{0.037}\\
\hline
\end{tabular}
}
\end{table}

\begin{table}[t]
\centering
\scriptsize
\setlength\tabcolsep{5pt}
\caption{Quantitative comparisons on the lung infection CT dataset. Top $2$ scores are highlighted in \textBC{mycellreda}{red} and \textBC{mycellblue}{blue}, respectively.  ``$\dagger$'' represents the medicine-specific method. 
  }
  \label{tab:COVID_comparison}
\resizebox{\columnwidth}{!}
{
\begin{tabular}{r|c||ccccccc}
\hline
\rowcolor{mygray}
Methods&Backbone & DSC $\uparrow$ & Precision $\uparrow$  &  Recall  $\uparrow$  & $S_{\alpha}$ $\uparrow$ & MAE $\downarrow$\\
\hline
U-Net~\cite{UNet} & R2-50 & 0.736 & 0.782 & 0.793 & 0.834 & \cellcolor{mycellblue}\textBC{white}{0.007} \\
U-Net++~\cite{UNet++} & R2-50& 0.592 & 0.637 & 0.748 & 0.806 & 0.010 \\
Atten-UNet~\cite{Attention_UNet} & R2-50& 0.650 & 0.755 & 0.715 & 0.801  & 0.010 \\
UTNet~\cite{UTNet} & R-50 + ViT-B16& 0.735 & 0.782 & 0.786 & 0.836  & \cellcolor{mycellblue}\textBC{white}{0.007} \\
TransUNet~\cite{TransUNet}  &R-50 + ViT-B16& 0.710 & 0.770 & 0.776 & 0.831  & \cellcolor{mycellblue}\textBC{white}{0.007} \\
Inf-Net$^{\dagger}$~\cite{Inf-Net} & R2-50& \cellcolor{mycellblue}\textBC{white}{0.783} & 0.774 & \cellcolor{mycellreda}\textBC{white}{0.852}& \cellcolor{mycellblue}\textBC{white}{0.853} &\cellcolor{mycellblue}\textBC{white}{0.007}\\
BCS-Net$^{\dagger}$~\cite{BCS-Net} & R2-50& 0.763 & 0.775 & 0.763 & 0.840 & \cellcolor{mycellblue}\textBC{white}{0.007}\\
\hline
MSNet~\cite{MSNet} & R2-50& 0.779 & \textBC{blue}{0.815} & 0.802 & 0.846 & \cellcolor{mycellblue}\textBC{white}{0.007}\\
{M$^{2}$SNet} & R2-50 & \cellcolor{mycellreda}\textBC{white}{0.795} & \cellcolor{mycellreda}\textBC{white}{0.825} & \cellcolor{mycellblue}\textBC{white}{0.813} & \cellcolor{mycellreda}\textBC{white}{0.855}  &\cellcolor{mycellreda}\textBC{white}{0.006}\\
\hline
\end{tabular}
}
\end{table}

\begin{table}[!t]
\centering
\scriptsize
\setlength\tabcolsep{3pt}
\caption{ Quantitative comparisons on the breast ultrasound dataset. Top $2$ scores are highlighted in \textBC{mycellreda}{red} and \textBC{mycellblue}{blue}, respectively.  ``$\dagger$'' represents the medicine-specific method. 
  }
  \label{tab:BUSI_comparison}
\resizebox{\columnwidth}{!}
{
\begin{tabular}{r|c||ccccc}
\hline
\rowcolor{mygray}
Methods & Backbone &Jaccard $\uparrow$ & Precision $\uparrow$  &  Recall  $\uparrow$  & Specificity $\uparrow$ &Dice $\uparrow$ \\
\hline
U-Net~\cite{UNet}  & R2-50& 51.54$\pm$1.17 & 56.84$\pm$1.85 & 71.18$\pm$2.87 & 95.20$\pm$0.83 & 57.61$\pm$1.26 \\
U-Net++~\cite{UNet++} & R2-50& 50.17$\pm$1.71 & 57.34$\pm$2.29 & 70.81$\pm$1.43 & 95.85$\pm$0.36 & 58.54$\pm$1.84 \\
Atten-UNet~\cite{Attention_UNet}  & R2-50& 46.91$\pm$4.40 & 55.84$\pm$3.81 & 70.77$\pm$2.41 & 95.56$\pm$0.62 & 57.39$\pm$2.78 \\
UTNet~\cite{UTNet}& R-50 + ViT-B16  & 67.46$\pm$1.78 & 79.88$\pm$1.22 & 74.82$\pm$1.95 & \cellcolor{mycellreda}\textBC{white}{98.37$\pm$0.36} & 74.41$\pm$1.39 \\
TransUNet~\cite{TransUNet}& R-50 + ViT-B16  & \cellcolor{mycellblue}\textBC{white}{71.47$\pm$0.98} & 81.66$\pm$1.52 & \cellcolor{mycellblue}\textBC{white}{80.78$\pm$1.63} & 98.05$\pm$0.25 & \cellcolor{mycellblue}\textBC{white}{79.00$\pm$0.79} \\
SKU-Net$^{\dagger}$~\cite{SKUNet}& SKs  & 64.48$\pm$2.37 & 75.37$\pm$3.22 & 78.56$\pm$3.27 & 97.21$\pm$1.02 & 74.03$\pm$2.21 \\
NU-Net$^{\dagger}$~\cite{NU-net}& Deeper U-Net& 68.86$\pm$1.99 & 78.90$\pm$2.26 & \cellcolor{mycellreda}\textBC{white}{82.48$\pm$2.14} & 97.79$\pm$0.87 & 77.79$\pm$1.88 \\
\hline
MSNet~\cite{MSNet} & R2-50 & 70.57$\pm$3.38 & \cellcolor{mycellblue}\textBC{white}{82.73$\pm$2.72} & 78.50$\pm$4.56 & 98.29$\pm$0.44 & 78.18$\pm$3.75 \\
{M$^{2}$SNet} & R2-50 & \cellcolor{mycellreda}\textBC{white}{71.52$\pm$1.19} &  \cellcolor{mycellreda}\textBC{white}{84.00$\pm$0.55} &{79.97$\pm$1.65} & \cellcolor{mycellblue}\textBC{white}{98.35$\pm$0.20} & \cellcolor{mycellreda}\textBC{white}{79.21$\pm$1.28} \\

\hline
\end{tabular}
}
\end{table}

\begin{table}[!t]
\centering
\scriptsize
\setlength\tabcolsep{2pt}
\caption{The leaderboard of MICCAI2022 Challenge: Glaucoma Oct Analysis and Layer Segmentation (GOALS). Top $2$ scores are highlighted in \textBC{mycellreda}{red} and \textBC{mycellblue}{blue}, respectively. 
  }
  \label{tab:OCT_layer_comparison}
\resizebox{\columnwidth}{!}
{
\begin{tabular}{r||cc|cc|cc}
\hline
\rowcolor{mygray}
  & \multicolumn{2}{c|}{RNFL} & \multicolumn{2}{c|}{GCIPL}& \multicolumn{2}{c}{Choroid} \\
 \rowcolor{mygray}
  Teams&Dice $\uparrow$ & MED $\downarrow$  & Dice  $\uparrow$  & MED $\downarrow$&Dice $\uparrow$ &MED $\downarrow$\\
\hline
LaTIM & 0.9547& 	1.1554& 	0.8950	& 1.3804	& 0.9557& 	1.7120\\
OPTIMA-MUW & 0.9559	& 1.1084& 	0.8955	& 1.2838& 	0.9550	& 1.7599 \\
WRMT & 0.9560	 & 1.1020	 &  \cellcolor{mycellblue}\textBC{white}{0.8980}&	1.2719 & 	0.9567 & 	1.6769\\
Miracle-boyi & 0.9562	 &1.1065	 &0.8904 &	1.3346 &	0.9542	 &1.7904 \\
SZUMed & 0.9555& 	1.1289& 	0.8921	&1.3565& 	0.9553& 1.7628\\
MedicalExplorer &0.9565&	1.0888&	0.8941&	1.2972&	0.9563&	1.7452\\
AUTOMATE  & 0.9561&	1.1014&	0.8966&	1.2848&	0.9569	&1.6767\\
SJMED  & 0.9565 &	1.0899	 &0.8970 &	1.2801	 & \cellcolor{mycellreda}\textBC{white}{0.9578} &	 \cellcolor{mycellblue}\textBC{white}{1.6456}\\
\rowcolor{mygray}
\textbf{IIAU-Segmentors (Ours)} &  \cellcolor{mycellreda}\textBC{white}{0.9576} & 	 \cellcolor{mycellreda}\textBC{white}{1.0827}	 & 0.8953	 &  \cellcolor{mycellreda}\textBC{white}{1.2322} & 	 \cellcolor{mycellreda}\textBC{white}{0.9578} & 	1.6468\\
Vision Wise &  \cellcolor{mycellblue}\textBC{white}{0.9569}	& \cellcolor{mycellblue}\textBC{white}{1.0833}	& \cellcolor{mycellreda}\textBC{white}{0.8992}& \cellcolor{mycellblue}\textBC{white}{1.2566}&	 \cellcolor{mycellblue}\textBC{white}{0.9576}	& \cellcolor{mycellreda}\textBC{white}{1.6295}\\
\hline
\end{tabular}
}
\end{table}

\begin{table}[!t]
  \centering
  \scriptsize
  \renewcommand{\arraystretch}{1.1}
  \setlength\tabcolsep{5pt}
  \caption{ The FLOPs, parameters and speed of different methods. The best and worst results are shown in \textBC{mycellreda}{red} and \textBC{mycellblue}{blue}, respectively. 
  }\label{tab:FLOPs_para}
  \resizebox{\columnwidth}{!}
  {
\begin{tabular}{r||c|c|c|c|c|c|c|c|c}
  \hline
  \rowcolor{mygray}
  Metrics & U-Net  & U-Net++ &Atten\_UNet & UTNet &TransUNet&PraNet&Inf-Net&BCSNet&M$^{2}$SNet\\
  \hline
   FLOPs (GB) $\downarrow$& $\sim$12.2  & \cellcolor{mycellblue}\textBC{white}{$\sim$39.3}  & $\sim$9.7 & $\sim$27.1 & $\sim$24.7 &12.0 &12.4 &14.8 &  \cellcolor{mycellreda}\textBC{white}{$\sim$9.0}  \\
  \hline
  Params (MB) $\downarrow$&  \cellcolor{mycellreda}\textBC{white}{$\sim$7.2}  & $\sim$19.2  & $\sim$12.6 & $\sim$34.0 & \cellcolor{mycellblue}\textBC{white}{$\sim$93.2}  & 30.5& 31.1& 44.8 & $\sim$27.7   \\
   \hline
  Speed (FPS) $\uparrow$&  \cellcolor{mycellreda}\textBC{white}{$\sim$111}  & $\sim$82  & $\sim$106 &\cellcolor{mycellblue}\textBC{white}{$\sim$44} & {$\sim$59}  &$\sim$74& $\sim$68& $\sim$49& $\sim$90   \\
  \hline
\end{tabular}
}
\end{table}

 \begin{table*}[!t]
  \centering
  \scriptsize
  \renewcommand{\arraystretch}{1.1}
  \setlength\tabcolsep{2pt}
    \caption{{Quantitative results at different training and inference codes frameworks. ``$^{\blacktriangle}$'' and ``$^{\blacktriangledown}$'' represent the model trained on the MSNet~\cite{MSNet} and nnU-Net~\cite{nnU-Net} code framework, respectively.}}\label{tab:ablation_study_2}
     \resizebox{\linewidth}{!}
     {
\begin{tabular}{r||cccc|cccc|ccc|ccc}
	 \hline
	 \rowcolor{mygray}
 & \multicolumn{4}{c|}{ETIS} & \multicolumn{4}{c|}{CVC-300-TV} & \multicolumn{3}{c|}{Lung Infection}& \multicolumn{3}{c}{BUSI}\\
	\rowcolor{mygray}
	 Methods &mDice  $\uparrow$ & mIoU  $\uparrow$ & $F_\beta^w$  $\uparrow$&$E_\phi^{max}$  $\uparrow$& mDice  $\uparrow$ & mIoU  $\uparrow$ & $S_{\alpha}$ $\uparrow$&MAE$\downarrow$& DSC $\uparrow$ & $S_{\alpha}$ $\uparrow$&MAE$\downarrow$& Jaccard $\uparrow$&Specificity $\uparrow$ & Dice $\uparrow$\\
       \hline
nnU-Net$^{\blacktriangledown}$ & {0.554} &{0.477} &0.510 &{0.730} & {0.742}&{0.670} &{0.848} &{0.016} &{0.819} &{0.854} & {0.005}&71.30$\pm$2.29&97.46$\pm$0.46&78.96$\pm$2.09\\
M$^{2}$SNet$^{\blacktriangle}$  & {0.749} &0.678 &{0.712} & {0.872}&{0.876} &{0.805} &{0.918} & {0.010}&{0.795} &{0.855} & {0.006}&71.52$\pm$1.19&98.35$\pm$0.20&79.21$\pm$1.38 \\
M$^{2}$SNet$^{\blacktriangledown}$   & {0.780} &0.711 &{0.754} & {0.899} &{0.897} &{0.832} &{0.930} & {0.009}&{0.848} &{0.855} & {0.005}&74.43$\pm$1.74&98.94$\pm$0.22&80.26$\pm$1.26\\

	\hline
	
	\end{tabular}
	}
\end{table*}

\subsection{Implementation Details}
Our model is implemented based on the PyTorch framework and trained on a single 2080Ti GPU with mini-batch size $16$.  We resize the inputs to $352 \times 352$ and employ a general multi-scale training strategy as most methods~\cite{F3Net,GCPANet,Rank-Net_COD,SPNet_RGBDSOD,PraNet,MSNet}. Random horizontally flipping and random rotate data augmentation are used to avoid overfitting. For the optimizer, we adopt the stochastic gradient descent (SGD). The momentum and weight decay are set as $0.9$ and $0.0005$, respectively.  Maximum learning rate is set to $0.005$ for backbone and $0.05$ for other parts. Warm-up and linear decay strategies  are  used to  adjust  the  learning  rate. 
For any medical image sub-tasks, the above training strategy is used for all the multi-scale subtraction models involved in this paper. 
The difference among these models is only in the number of training epochs due to different convergence speeds. Specifically, the number of training epochs settings in the  polyp segmentation, lung infection, breast tumor segmentation and OCT layer segmentation are $50$, $200$, $100$ and $100$, respectively.

\begin{figure*}[t]
\centering
\includegraphics[width=\linewidth]{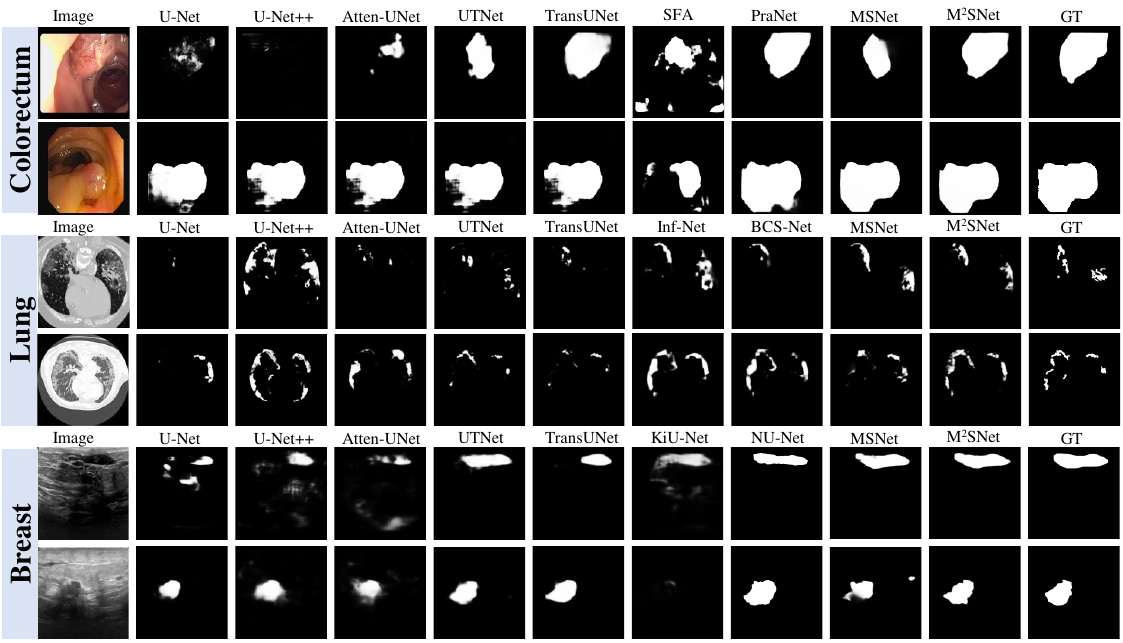}
\caption{Visual comparison of different medicine-general and medicine-specific methods.}
\label{fig:visual_comparison}
\end{figure*}

\subsection{Comparisons with Medicine-general and Medicine-specific Methods}
For a fair comparison, we compare not only with medicine-specific methods but also with representative medicine-general methods, including UNet~\cite{UNet}, UNet++~\cite{UNet++}, Attention U-Net~\cite{Attention_UNet}, UTNet~\cite{UTNet} and TransUNet~\cite{TransUNet}. Based on the open-source codes, we retrain these medicine-general methods on the same training sets as our models. 
\\
\noindent$\bullet$ In Tab.~\ref{tab:image_polyp_comparison}, among $30$ scores of all \textit{\textbf{image polyp}} datasets, our multi-scale subtraction models (MSNet + M$^{2}$SNet) achieve the best performance in terms of all six metrics. The M$^{2}$SNet even outperforms the video-based polyp segmentation method PNS-Net~\cite{PNSNet} on the \textit{\textbf{video polyp}} datasets, as shown in Tab.~\ref{tab:video_polyp_comparison}.
\\
\noindent$\bullet$ 
Tab.~\ref{tab:COVID_comparison} shows performance comparisons on the \textit{\textbf{lung infection CT}} datasets. Compared to the second best method (Inf-Net~\cite{Inf-Net}), M$^{2}$SNet achieves an important improvement of $1.5\%$, $6.6\%$ and $14.3\%$ in terms of DSC, Precision and MAE, respectively.
\\
\noindent$\bullet$ 
Tab.~\ref{tab:BUSI_comparison} shows performance comparisons with \textit{\textbf{breast tumor segmentation}} methods. Following most methods~\cite{SKUNet,NU-net} in this field, we adopt the four-fold cross-validation strategy. 
M$^{2}$SNet achieves the best performance in terms of the Jaccard, Precision and Dice metrics, which outperforms representative transformer-based methods, UTNet and TransUNet. Moreover, M$^{2}$SNet has the smallest mean standard deviation ($0.97$) under the five metrics, which indicates its performance stability.
\\
\noindent$\bullet$ 
{Generally speaking, different training and inference frameworks will produce different final performance. For a fair comparison, we train two versions of M$^{2}$SNet, one version follows the MSNet~\cite{MSNet} framework and the other based on the popular nnU-Net~\cite{nnU-Net} framework. Benefiting from the nnU-Net framework with many effective tricks, the performance of  M$^{2}$SNet can be further improved, as shown in Tab.~\ref{tab:ablation_study_2}.}
\\
\noindent$\bullet$ 
In Tab.~\ref{tab:OCT_layer_comparison}, we list the top $10$ \textit{\textbf{OCT layer segmentation}} results in MICCAI2022 GOALS Challenge. 
Based on the M$^{2}$SNet, we won the second place (2/100) according to the weighted score of six results in three different layers. It is worth noting that our method ranks the top $1$ in four out of six metrics.
\\
\noindent$\bullet$ 
As can be seen from Tab.~\ref{tab:image_polyp_comparison} - Tab.~\ref{tab:BUSI_comparison}, our M$^{2}$SNet consistently surpasses other medicine-general methods in all medical segmentation sub-branches. In Tab.~\ref{tab:FLOPs_para}, we list the FLOPs and parameters of different medicine-general methods. It can be seen that our method has only 9GB in FLOPs, which has obvious advantages in terms of computational efficiency.
\\
\noindent$\bullet$ 
{In Tab.~\ref{tab:FLOPs_para}, we compare the model efficiency in terms of FLOPs, parameters and inference speed. It can be seen that our method ranks first in terms of FLOPs, which has obvious advantages among both medicine-general and medicine-specific methods. As can be seen from Tab.~\ref{tab:image_polyp_comparison} - Tab.~\ref{tab:BUSI_comparison}, our M$^{2}$SNet consistently surpasses other methods in all medical segmentation sub-branches. Therefore, the proposed M$^{2}$SNet achieves good balance on accuracy and efficiency.}
\\
\noindent$\bullet$ 
Fig.~\ref{fig:visual_comparison} depicts a qualitative comparison with other methods. It can be seen that the results of M$^{2}$SNet have greater advantages in terms of detection accuracy, completeness, and sharpness across different image modalities.

\begin{table}[t]
  \centering
  \scriptsize
  \renewcommand{\arraystretch}{1.1}
  \setlength\tabcolsep{2pt}
    \caption{{Quantitative comparison of  medicine-specific methods on
four different medical lesion segmentation tasks in terms of ASD and HD95.}}\label{tab:comparison_asd_hd95}
     \resizebox{\linewidth}{!}
     {
	\begin{tabular}{r||cc|cc|cc|cc}
	 \hline
	 \rowcolor{mygray}
   &\multicolumn{2}{c|}{ETIS} & \multicolumn{2}{c|}{CVC-300-TV} & \multicolumn{2}{c|}{Lung Infection}& \multicolumn{2}{c}{BUSI}\\
	\rowcolor{mygray}
	 Methods 
     &ASD(mm)  $\downarrow$ & HD95(mm)  $\downarrow$ 
     &ASD(mm)  $\downarrow$ & HD95(mm)  $\downarrow$  &ASD(mm)  $\downarrow$ & HD95(mm)   $\downarrow$ 
     &ASD(mm)  $\downarrow$ & HD95(mm)  $\downarrow$\\
	\hline
PraNet  
& {1.154} &2.230  
& {-} &-
& {-} &-
& {-} &-
\\
PNS-Net  
& {-} &-  
& {1.726} &3.074
& {-} &-
& {-} &-
\\
Inf-Net 
& {-} &-  
& {-} &-
& {4.589} &9.845
& {-} &-
\\
NU-Net  
& {-} &-  
& {-} &-
& {-} &-
& {3.576} &8.745
\\
\hline
MSNet  
& {0.602} &1.544  
& {0.904} &1.996
& {2.322} &8.074
& {1.967} &7.024
\\
M$^{2}$SNet 
& \cellcolor{mycellreda}\textBC{white}{0.584} &\cellcolor{mycellreda}\textBC{white}{1.280}  
& \cellcolor{mycellreda}\textBC{white}{0.847} &\cellcolor{mycellreda}\textBC{white}{1.874}
& \cellcolor{mycellreda}\textBC{white}{2.021} &\cellcolor{mycellreda}\textBC{white}{7.412}
& \cellcolor{mycellreda}\textBC{white}{1.819} &\cellcolor{mycellreda}\textBC{white}{6.451}
\\
\hline	
	\end{tabular}
	}
\end{table}

\begin{table}[t]
\centering
\scriptsize
\setlength\tabcolsep{2pt}
\caption{{Quantitative comparison of unified, generalist and Mamba-based models on three different medical lesion segmentation tasks. The best results are shown in \textBC{mycellreda}{red}.}  
  }
  \label{tab:comparison_unified_model}
\resizebox{\columnwidth}{!}
{
\begin{tabular}{r|c||cc|cc|cc}
\hline
\rowcolor{mygray}
 & & \multicolumn{2}{c|}{Lung Infection} & \multicolumn{2}{c|}{Breast Lesion}& \multicolumn{2}{c}{Polyp} \\
 \rowcolor{mygray}
  Methods&Backbone&Dice $\uparrow$ & mIoU $\uparrow$  & Dice  $\uparrow$  & mIoU $\uparrow$  &Dice $\uparrow$ & mIoU $\uparrow$  \\
\hline

SegMamba~\cite{Segmamba} &Customized Design& 
0.680	 & 0.802	 &  {0.827}&	0.848 & 	0.815 & 0.848\\
H-vmunet~\cite{H-vmunet}  &Customized Design&
0.684& 	0.809& 	0.818	& 0.840	& 0.819& 	0.851\\
UniverSeg~\cite{UniMRSeg} &ResNet-101~\cite{ResNet}& 
0.673& 	0.368& 	0.775	& 0.600	& 0.553& 	0.261\\
SegGPT~\cite{SegGPT} & ViT-L~\cite{ViT}& 
0.131	& 0.553& 	0.336	& 0.603& 	0.568	& 0.707 \\
MaskSAM~\cite{MaskSAM} & ViT-B~\cite{ViT}& 
0.698	& 0.810& 	0.835	& 0.851& 	0.810	& 0.858 \\
SAM 2~\cite{SAM2} & Hiera-L~\cite{Hiera}& 
0.382	 &0.655	 &0.539 &	0.712 &	0.499	 &0.706 \\
MedSAM-2~\cite{MedSAM-2}  & Hiera-L~\cite{Hiera}& 
0.512	 &0.704	 &0.787 &	0.802 &	0.765	 &0.811 \\
Spider~\cite{Spider}  &ConvNeXt-B~\cite{ConvNeXt}& 
0.696	 & 0.813	 &  {0.838}&	0.866 & 	0.824 & 	\cellcolor{mycellreda}\textBC{white}{0.866}\\
{M$^{2}$SNet} & ConvNeXt-B~\cite{ConvNeXt} & \cellcolor{mycellreda}\textBC{white}{0.702} & \cellcolor{mycellreda}\textBC{white}{0.818} & \cellcolor{mycellreda}\textBC{white}{0.844} & \cellcolor{mycellreda}\textBC{white}{0.870}  &\cellcolor{mycellreda}\textBC{white}{0.832}&\cellcolor{mycellreda}\textBC{white}{0.866}\\

\hline
\end{tabular}
}
\end{table}

\subsection{{Evaluation on Boundary Accuracy}}
{
In addition to overlap-based metrics such as Dice and IoU, distance-based metrics are widely used to evaluate boundary accuracy in medical image segmentation. In particular, the Average Surface Distance (ASD) measures the average symmetric distance between predicted and ground-truth surfaces, while the 95th percentile Hausdorff Distance (HD95) captures the worst-case boundary discrepancy under a robust percentile criterion. Compared with region-overlap metrics, these indicators provide more direct assessment of contour precision and structural consistency.
We conduct comparison with medicine-specific methods across four different tasks.  
As shown in Tab.~\ref{tab:comparison_asd_hd95}, M$^{2}$SNet consistently achieves the lowest ASD and HD95 scores across all tasks, indicating more accurate boundary delineation and reduced extreme surface deviations. These improvements further validate the effectiveness of the proposed multi-scale subtraction design in preserving fine-grained structural information. Importantly, the superiority observed under overlap-based metrics remains consistent when evaluated using distance-based criteria, demonstrating the robustness and reliability of the proposed framework.
}

\subsection{Comparisons with Unified, Generalist and Mamba-based Methods}
In recent years, with the rapid development of large‐scale and foundation models, there has been a growing interest in building \emph{unified} and \emph{generalist} models that can solve multiple tasks with a single set of model parameters.  
These models~\cite{UniverSeg,SegGPT,SAM2} aim to break the traditional paradigm of training task‐specific networks by introducing shared representations, prompt mechanisms, or in‐context learning strategies so that one model can flexibly adapt to diverse downstream tasks.  
Following the evaluation protocol of SAM-Eva~\cite{SAM-Eva}, we evaluate our proposed M$^{2}$SNet against representative unified and generalist methods on three challenging cross‐modality, cross‐lesion segmentation tasks, including lung infection, breast lesion, and polyp segmentation.  

As shown in Tab.~\ref{tab:comparison_unified_model}, UniverSeg~\cite{UniverSeg}, SegGPT~\cite{SegGPT}, and SAM~2~\cite{SAM2} exhibit clear performance drops when directly applied to cross‐modality and cross‐lesion scenarios. These models mainly rely on prompt embeddings or in‐context examples to handle unseen tasks but still struggle with the strong context‐dependence and heterogeneous appearances in medical images.  
Spider~\cite{Spider}, a unified context‐dependent segmentation model, achieves better performance than the above generalist models, demonstrating the effectiveness of high‐level concept matching mechanisms.
Our M$^{2}$SNet consistently surpasses all the competitors across all three tasks, obtaining the highest Dice and mIoU scores. In our experiments, we jointly train M$^{2}$SNet on the training sets of the three tasks. Thanks to the implicit prompts naturally embedded in modality and lesion types, M$^{2}$SNet can perform joint learning of all data under a single parameter set, thus simultaneously achieving unified modelling and superior performance. By contrast, other methods are trained with a much broader range of tasks and domains but still underperform on these specific cross‐modality, cross‐lesion settings.  
Although the scope of M$^{2}$SNet is narrower than that of some generalist models, the results provide an important insight: future universal and unified models can be organized into multiple levels or hierarchies of unification to better balance task coverage and performance for real‐world applications.

{
In addition to unified frameworks, we further compare with recent Mamba-based and SAM-based medical segmentation methods. Specifically, SegMamba~\cite{Segmamba} and H-vmunet~\cite{H-vmunet} represent state-space–based architectures tailored for medical imaging, while MaskSAM~\cite{MaskSAM} and MedSAM-2~\cite{MedSAM-2} adapt foundation segmentation models to the medical domain. As shown in Tab.~\ref{tab:comparison_unified_model}, although SegMamba and H-vmunet demonstrate competitive performance compared with some unified generalist models, their overall results remain consistently lower than those of M$^{2}$SNet across all three tasks and evaluation metrics. This indicates that while state-space modeling improves long-range dependency capture and computational efficiency, it does not inherently guarantee superior lesion discrimination capability in cross-modality scenarios.
Similarly, MaskSAM and MedSAM-2 benefit from strong pre-training on large-scale data and exhibit relatively stable performance. However, their segmentation accuracy is still inferior to M$^{2}$SNet, especially in scenarios involving subtle boundary variations and heterogeneous lesion appearances. This suggests that directly adapting foundation segmentation models to medical imaging, even with domain-specific fine-tuning, may not fully address the strong context dependence and fine-grained structural requirements of lesion segmentation.
From a modeling perspective, Mamba-based architectures primarily emphasize efficient sequence modeling and long-range dependency propagation. However, medical lesion segmentation often relies more heavily on precise boundary modeling, multi-scale structural aggregation, and strong local inductive bias. Pure state-space modeling lacks the explicit spatial hierarchy and strong locality priors that are naturally embedded in convolutional designs. Although hybrid variants partially alleviate this limitation, their representation learning capacity for fine-grained lesion structures remains constrained.
As a result, CNN-based and Transformer-based architectures continue to dominate mainstream medical image segmentation research.
}

 \begin{table}[!t]
  \centering
  \scriptsize
  \renewcommand{\arraystretch}{1.1}
  \setlength\tabcolsep{1pt}
    \caption{Ablation experiments of the subtraction unit, inter-layer multi-scale aggregation and LossNet.}\label{tab:ablation_study_1}
     \resizebox{\columnwidth}{!}
     {
\begin{tabular}{r||cccc|cccc}
	 \hline
	 \rowcolor{mygray}
	 & \multicolumn{4}{c|}{ColonDB} & \multicolumn{4}{c}{ETIS} \\
	\rowcolor{mygray}
	 Methods & mDice  $\uparrow$  & mIoU  $\uparrow$ & $F_\beta^w$  $\uparrow$&$E_\phi^{max}$  $\uparrow$ & mDice  $\uparrow$ & mIoU   $\uparrow$& $F_\beta^w$  $\uparrow$ &$E_\phi^{max}$  $\uparrow$ \\
	\hline
	baseline ($SU_{1}^{i}$) &{0.678} &{0.607} &{0.659} & {0.825} & {0.588} &0.549 &{0.532} & {0.707} \\
	\hline
	+ $SU_{2}^{i}$ &{0.731} &{0.652} &{0.703} & {0.861} & {0.642} &0.579 &{0.586} & {0.745} \\
	\hline
	+ $SU_{3}^{i}$ &{0.733} &{0.659} &{0.712} & {0.861} & {0.642} &0.580 &{0.581} & {0.745} \\
	\hline
	+ $SU_{4}^{i}$ &{0.750} &{0.676} &{0.729} & {0.872} & {0.643} &0.580 &{0.585} & {0.757} \\
	\hline
	+ $SU_{5}^{i}$ &{0.749} &{0.676} &{0.729} & {0.878} & {0.643} &0.582 &{0.600} & {0.787} \\
	\hline
	+ $\mathcal{L}_{f}$ &{0.755} &{0.678} &{0.737} & {0.883} & {0.719} &0.664 &{0.678} & {0.830} \\
	\hline
	$SU$ $\rightarrow$ $AU$ &{0.697} &{0.630} &{0.676} & {0.839} & {0.680} &0.621 &{0.636} & {0.820} \\
	\hline
	
	\end{tabular}
	}
\end{table}

\begin{table*}[!t]
  \centering
  \scriptsize
  \renewcommand{\arraystretch}{1.1}
  \setlength\tabcolsep{2pt}
    \caption{{{Ablation experiments of the loss functions.}} }\label{tab:ablation_study_3}
     \resizebox{\linewidth}{!}
     {
\begin{tabular}{r||cccc|cccc|ccc|ccc}
	 \hline
	 \rowcolor{mygray}
 & \multicolumn{4}{c|}{ETIS} & \multicolumn{4}{c|}{CVC-300-TV} & \multicolumn{3}{c|}{Lung Infection}& \multicolumn{3}{c}{BUSI}\\
	\rowcolor{mygray}
	 Methods &mDice  $\uparrow$ & mIoU  $\uparrow$ & $F_\beta^w$  $\uparrow$&$E_\phi^{max}$  $\uparrow$& mDice  $\uparrow$ & mIoU  $\uparrow$ & $S_{\alpha}$ $\uparrow$&MAE$\downarrow$& DSC $\uparrow$ & $S_{\alpha}$ $\uparrow$&MAE$\downarrow$& Jaccard $\uparrow$&Specificity $\uparrow$ & Dice $\uparrow$\\
       \hline
M$^{2}$SNet (LossNet)   & {0.749} &0.678 &{0.712} & {0.872}&{0.876} &{0.805} &{0.918} & {0.010}&{0.795} &{0.855} & {0.006}&71.52$\pm$1.19&98.35$\pm$0.20&79.21$\pm$1.38 \\
M$^{2}$SNet (w/o LossNet)  & {0.725} &0.668 &{0.690} & {0.854} &{0.856} &{0.783} &{0.905} & {0.012}&{0.783} &{0.850} & {0.007}&71.22$\pm$1.90&98.28$\pm$0.22&79.02$\pm$1.94\\
M$^{2}$SNet (w/o $\mathcal{L}_{IoU}^w \& \mathcal{L}_{BCE}^w$) & {0.700} &0.649 &{0.678} & {0.836} &{0.830} &{0.762} &{0.895} & {0.013}&{0.771} &{0.838} & {0.007}&70.82$\pm$1.85&98.01$\pm$0.25&78.66$\pm$1.97\\
M$^{2}$SNet (w/o LossNet + Deep Supervision) & {0.722} &0.662 &{0.688} & {0.852} &{0.860} &{0.784} &{0.903} & {0.012}&{0.786} &{0.853} & {0.007}&71.15$\pm$2.07&98.14$\pm$0.26&78.97$\pm$1.97\\
M$^{2}$SNet (LossNet + Deep Supervision) & {0.753} &0.681 &{0.717} & {0.875}&{0.879} &{0.811} &{0.921} & {0.009}&{0.797} &{0.859} & {0.006}&71.56$\pm$1.22&98.34$\pm$0.21&79.24$\pm$1.46 \\

	\hline
	
	\end{tabular}
	}
\end{table*}

\begin{table*}[t]
  \centering
  \scriptsize
  \renewcommand{\arraystretch}{1.1}
  \setlength\tabcolsep{2pt}
    \caption{{Ablation experiments on different pre-trained networks used as LossNet.}}\label{tab:ablation_study_lossnet_different_backbone}
     \resizebox{\linewidth}{!}
     {
\begin{tabular}{r||cccc|cccc|ccc|ccc}
	 \hline
	 \rowcolor{mygray}
 & \multicolumn{4}{c|}{ETIS} & \multicolumn{4}{c|}{CVC-300-TV} & \multicolumn{3}{c|}{Lung Infection}& \multicolumn{3}{c}{BUSI}\\
	\rowcolor{mygray}
	 Methods &mDice  $\uparrow$ & mIoU  $\uparrow$ & $F_\beta^w$  $\uparrow$&$E_\phi^{max}$  $\uparrow$& mDice  $\uparrow$ & mIoU  $\uparrow$ & $S_{\alpha}$ $\uparrow$&MAE$\downarrow$& DSC $\uparrow$ & $S_{\alpha}$ $\uparrow$&MAE$\downarrow$& Jaccard $\uparrow$&Specificity $\uparrow$ & Dice $\uparrow$\\
       \hline
VGG-16   & {0.749} &0.678 &{0.712} & {0.872}&{0.876} &{0.805} &{0.918} & {0.010}&{0.795} &{0.855} & {0.006}&71.52$\pm$1.19&98.35$\pm$0.20&79.21$\pm$1.38 \\

ResNet-50  & {0.737} &0.671 &{0.699} & {0.858} &{0.861} &{0.794} &{0.910} & {0.012}&{0.788} &{0.852} & {0.007}&71.28$\pm$1.48&98.27$\pm$0.26&79.08$\pm$1.56\\

Swin-T & {0.729} &0.660 &{0.685} & {0.850} &{0.854} &{0.789} &{0.907} & {0.012}&{0.784} &{0.850} & {0.007}&71.10$\pm$1.56&98.20$\pm$0.29&78.99$\pm$1.84\\
	\hline
	
	\end{tabular}
	}
\end{table*}

 \begin{table*}[!t]
  \centering
  \scriptsize
  \renewcommand{\arraystretch}{1.1}
  \setlength\tabcolsep{2pt}
    \caption{{Ablation experiments of the kernel scales and weights in intra-layer multi-scale subtraction.}}\label{tab:ablation_study_4}
     \resizebox{\linewidth}{!}
     {
\begin{tabular}{r||cccc|cccc|ccc|ccc}
	 \hline
	 \rowcolor{mygray}
 & \multicolumn{4}{c|}{ETIS} & \multicolumn{4}{c|}{CVC-300-TV} & \multicolumn{3}{c|}{Lung Infection}& \multicolumn{3}{c}{BUSI}\\
	\rowcolor{mygray}
	 Methods &mDice  $\uparrow$ & mIoU  $\uparrow$ & $F_\beta^w$  $\uparrow$&$E_\phi^{max}$  $\uparrow$& mDice  $\uparrow$ & mIoU  $\uparrow$ & $S_{\alpha}$ $\uparrow$&MAE$\downarrow$& DSC $\uparrow$ & $S_{\alpha}$ $\uparrow$&MAE$\downarrow$& Jaccard $\uparrow$&Specificity $\uparrow$ & Dice $\uparrow$\\
       \hline
M$^{2}$SNet (Kernel Scales = 1)   & {0.719} &0.664 &{0.678} & {0.830}&{0.837} &{0.755} &{0.895} & {0.014} &{0.779} &{0.846} & {0.007}&70.57$\pm$3.38&98.29$\pm$0.44&78.18$\pm$3.75\\
M$^{2}$SNet (Kernel Scales = 1,3)  & {0.738} &0.672 &{0.695} & {0.858}&{0.862} &{0.787} &{0.906} & {0.012} &{0.789} &{0.850} & {0.007}&71.08$\pm$1.43&98.33$\pm$0.51&78.88$\pm$1.89\\
M$^{2}$SNet (Kernel Scales = 1,3,5)  & {0.749} &0.678 &{0.712} & {0.872}&{0.876} &{0.805} &{0.918} & {0.010}&{0.795} &{0.855} & {0.006}&71.52$\pm$1.19&98.35$\pm$0.20&79.21$\pm$1.38\\
M$^{2}$SNet (Kernel Scales = 1,3,7) & {0.745} &0.674 &{0.711} & {0.867}&{0.870} &{0.804} &{0.912} & {0.011} &{0.789} &{0.852} & {0.006}&71.48$\pm$1.25&98.30$\pm$0.19&79.17$\pm$1.44 \\
M$^{2}$SNet (Kernel Scales = 1,5,9)  & {0.737} &0.670 &{0.693} & {0.859}&{0.860} &{0.784} &{0.901} & {0.013} &{0.784} &{0.847} & {0.007}&71.01$\pm$1.54&98.28$\pm$0.55&78.89$\pm$1.93\\
M$^{2}$SNet (Kernel Scales = 1,3,5,7) & {0.748} &0.678 &{0.713} & {0.868}&{0.875} &{0.808} &{0.921} & {0.010} &{0.799} &{0.861} & {0.007}&71.56$\pm$1.18&98.40$\pm$0.21&79.25$\pm$1.36 \\
M$^{2}$SNet (Average Kernel $\rightarrow$ Gaussian Kernel)  & {0.670} &0.605 &{0.650} & {0.817} &{0.827} &{0.750} &{0.886} & {0.014}&{0.753} &{0.830} & {0.007}&68.18$\pm$2.34&97.74$\pm$0.83&77.52$\pm$2.42\\
	\hline
	
	\end{tabular}
	}
\end{table*}

\subsection{Ablation Study}
We take the common FPN network as the baseline to analyze the contribution of each component. 
\subsubsection{Effectiveness of the subtraction unit, inter-layer multi-scale subtraction aggregation}
The results are shown in Tab.~\ref{tab:ablation_study_1}. These defined feature subscripts are the same as those in Fig~\ref{fig:Pipeline}. 
First, we apply the basic subtraction unit (SU) to the baseline to get a  series of $SU_{2}^{i}$  features to participate in the feature aggregation calculated by Equ.~\ref{equ:1}. The gap between the `` + $SU_{2}^{i}$ ''  and the baseline demonstrates the effectiveness of the SU.  It can be seen that the usage of SU has a significant improvement on the ColonDB dataset compared to the baseline, with the gain of 7.8\%, 7.4\%, 6.7\% and 4.4\% in terms of mDice, mIoU, $F_\beta^w$, and $E_\phi^{max}$, respectively. Next, we gradually add  $SU_{3}^{i}$, $SU_{4}^{i}$ and $MS_{5}^{i}$ to achieve inter-layer multi-scale aggregation. The gap between the `` + $SU_{5}^{i}$ ''  and the `` + $SU_{2}^{i}$ '' quantitatively demonstrates the effectiveness of inter-layer multi-scale subtraction strategy. Next, we evaluate the benefit of $\mathcal{L}_{f}$.  Compared to the  `` + $SU_{5}^{i}$ '' model,  the  `` + $\mathcal{L}_{f}$ '' achieves significant performance improvement on the ETIS dataset, with the gain of 11.8\%, 14.1\%, 13.0\% and 5.5\% in terms of mDice, mIoU, $F_\beta^w$, and $E_\phi^{max}$, respectively. Besides,  we replace all subtraction units with the element-wise addition units (AU) and compare their performance. It can be seen that our subtraction units have significant advantage and no additional parameters are introduced.

\begin{figure*}[!t]
\centering
\includegraphics[width=\linewidth]{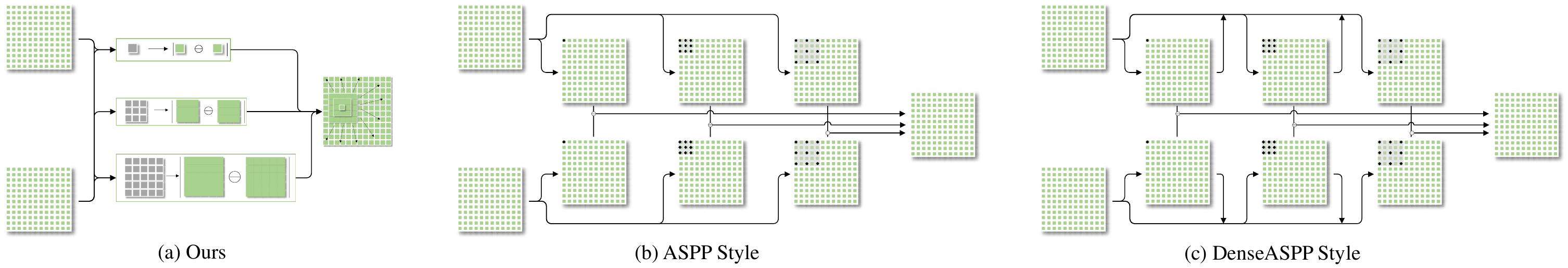}
\caption{Illustration of different multi-scale modules applied in the subtraction unit.}
\label{fig:aspp_denseaspp_ours_structure}
\end{figure*}

 \begin{table*}[!t]
  \centering
  \scriptsize
  \renewcommand{\arraystretch}{1.1}
  \setlength\tabcolsep{2pt}
    \caption{Quantitative comparisons of different multi-scale styles in intra-layer multi-scale subtraction design. Positive and negative gains are highlighted in \textBC{mycellreda}{red} and \textBC{mycellblue}{blue}, respectively. }\label{tab:ablation_study_5}
     \resizebox{\linewidth}{!}
     {
	\begin{tabular}{r||cc|cccc|cccc|ccc|ccc}
	 \hline
	 \rowcolor{mygray}
  & \multicolumn{2}{c|}{Efficiency}& \multicolumn{4}{c|}{ETIS} & \multicolumn{4}{c|}{CVC-300-TV} & \multicolumn{3}{c|}{Lung Infection}& \multicolumn{3}{c}{BUSI}\\
	\rowcolor{mygray}
	 Methods & FLOPs  $\downarrow$&Params  $\downarrow$&mDice  $\uparrow$ & mIoU  $\uparrow$ & $F_\beta^w$  $\uparrow$&$E_\phi^{max}$  $\uparrow$& mDice  $\uparrow$ & mIoU  $\uparrow$ & $S_{\alpha}$ $\uparrow$&MAE$\downarrow$& DSC $\uparrow$ & $S_{\alpha}$ $\uparrow$&MAE$\downarrow$& Jaccard $\uparrow$&Specificity $\uparrow$ & Dice $\uparrow$\\
	\hline
	MSNet &{9.0 G} &{27.7 M}  & {0.719} &0.664 &{0.678} & {0.830}&{0.837} &{0.755} &{0.895} & {0.014} &{0.779} &{0.846} & {0.007}&70.57$\pm$3.38&98.29$\pm$0.44&78.18$\pm$3.75\\
        \hline
M$^{2}$SNet (ASPP) &{13.8 G} &{29.2 M}  & {0.711} &0.644 &{0.671} & {0.838} &{0.827} &{0.752} &{0.890} & {0.014}&{0.809} &{0.855} & {0.006}&60.23$\pm$1.94&98.14$\pm$0.42&68.52$\pm$1.82\\
$\Delta$ gains &\cellcolor{mycellblue}\textBC{white}{- 53.3\%}&\cellcolor{mycellblue}\textBC{white}{- 5.4\%}& \cellcolor{mycellblue}\textBC{white}{-1.1\%}&\cellcolor{mycellblue}\textBC{white}{- 3.0\%}&\cellcolor{mycellblue}\textBC{white}{- 1.0\%}&\cellcolor{mycellblue}\textBC{white}{- 1.0\%}&\cellcolor{mycellblue}\textBC{white}{- 1.2\%}&\cellcolor{mycellblue}\textBC{white}{- 0.4\%}& \cellcolor{mycellblue}\textBC{white}{- 0.6\%}&\cellcolor{mycellreda}\textBC{white}{ 0.0\%}&\cellcolor{mycellreda}\textBC{white}{+ 3.9\%}&\cellcolor{mycellreda}\textBC{white}{+ 1.1\%}&\cellcolor{mycellreda}\textBC{white}{+ 14.3\%}&\cellcolor{mycellblue}\textBC{white}{- 14.7\%}&\cellcolor{mycellblue}\textBC{white}{- 0.2\%}&\cellcolor{mycellblue}\textBC{white}{- 12.4\%}\\
	\hline
M$^{2}$SNet (DenseASPP) &{17.5 G} &{29.6 M}  & {0.709} &0.645 &{0.668} & {0.833} &{0.869} &{0.798} &{0.916} & {0.010}&{0.798} &{0.853} & {0.006}&63.11$\pm$4.63&98.42$\pm$0.59&71.71$\pm$4.33\\
$\Delta$ gains &\cellcolor{mycellblue}\textBC{white}{- 94.4\%}&\cellcolor{mycellblue}\textBC{white}{- 6.9\%}& \cellcolor{mycellblue}\textBC{white}{-1.4\%}&\cellcolor{mycellblue}\textBC{white}{- 2.9\%}&\cellcolor{mycellblue}\textBC{white}{- 1.5\%}&\cellcolor{mycellblue}\textBC{white}{- 0.4\%}&\cellcolor{mycellreda}\textBC{white}{+ 3.8\%}&\cellcolor{mycellreda}\textBC{white}{+ 5.7\%}& \cellcolor{mycellreda}\textBC{white}{+ 2.3\%}&\cellcolor{mycellreda}\textBC{white}{+ 28.6\%}&\cellcolor{mycellreda}\textBC{white}{+ 2.4\%}&\cellcolor{mycellreda}\textBC{white}{+ 0.8\%}&\cellcolor{mycellreda}\textBC{white}{+ 14.3\%}&\cellcolor{mycellblue}\textBC{white}{- 10.6\%}&\cellcolor{mycellreda}\textBC{white}{+ 0.1\%}&\cellcolor{mycellblue}\textBC{white}{- 8.3\%}\\
	\hline
M$^{2}$SNet (Ours)&{9.0 G} &{27.7 M}  & {0.749} &0.678 &{0.712} & {0.872}&{0.876} &{0.805} &{0.918} & {0.010}&{0.795} &{0.855} & {0.006}&71.52$\pm$1.19&98.35$\pm$0.20&79.21$\pm$1.38 \\
$\Delta$ gains &\cellcolor{mycellreda}\textBC{white}{ 0.0\%}&\cellcolor{mycellreda}\textBC{white}{ 0.0\%}& \cellcolor{mycellreda}\textBC{white}{+ 4.2\%}&\cellcolor{mycellreda}\textBC{white}{+ 2.1\%}&\cellcolor{mycellreda}\textBC{white}{+ 5.0\%}&\cellcolor{mycellreda}\textBC{white}{+ 5.1\%}&\cellcolor{mycellreda}\textBC{white}{+ 4.7\%}&\cellcolor{mycellreda}\textBC{white}{+ 6.6\%}& \cellcolor{mycellreda}\textBC{white}{+ 2.6\%}&\cellcolor{mycellreda}\textBC{white}{+ 28.6\%}&\cellcolor{mycellreda}\textBC{white}{+ 2.1\%}&\cellcolor{mycellreda}\textBC{white}{+ 1.1\%}&\cellcolor{mycellreda}\textBC{white}{+ 14.3\%}&\cellcolor{mycellreda}\textBC{white}{+ 1.3\%}&\cellcolor{mycellreda}\textBC{white}{+ 0.1\%}&\cellcolor{mycellreda}\textBC{white}{+ 1.3\%}\\

	\hline
	
	\end{tabular}
	}
\end{table*}

\subsubsection{Effectiveness of loss function}
{In Tab.~\ref{tab:ablation_study_3}, we thoroughly verify the effectiveness of the loss function used in M$^{2}$SNet. The gap between M$^{2}$SNet (LossNet) and M$^{2}$SNet (w/o LossNet) demonstrates the general effectiveness of LossNet for different medical segmentation tasks. At the same time, without the auxiliary of LossNet, the performance of M$^{2}$SNet itself is good enough to surpass most of the methods in Tab.~\ref{tab:image_polyp_comparison} - Tab.~\ref{tab:BUSI_comparison}. 
The gap between M$^{2}$SNet (LossNet) and M$^{2}$SNet (w/o  $\mathcal{L}_{IoU}^w$ \& $\mathcal{L}_{BCE}^w$) shows the necessity of the weighted IoU and BCE loss, which provide a basic foreground region guidance for LossNet to focus on supervising multi-level lesion regions without distracting in the background area. }

{
We further verify the effectiveness of LossNet from the perspective of supervision strategy. Applying losses at different decoder scales, commonly referred to as deep supervision, mainly enforces pixel-level consistency between intermediate predictions and ground-truth masks. Although deep supervision can accelerate convergence and stabilize gradient propagation, it operates directly in the prediction space and typically relies on carefully designed task-specific loss functions.
In contrast, LossNet introduces supervision in a representation space rather than directly in the prediction space. By leveraging a VGG-16 network pre-trained on ImageNet, LossNet extracts hierarchical feature maps that encode rich visual cues, including fine-grained textures, structural patterns, and global semantic context. These multi-level perceptual representations provide comprehensive guidance, encouraging the predicted masks to match the structural and semantic characteristics of the ground truth from a higher-level perspective.
Moreover, the rich representations extracted by the pre-trained VGG-16 allow LossNet to employ a simple L2 loss for supervision, reducing dependence on manually designed segmentation-specific losses. To further examine the relationship between LossNet and deep supervision, we conduct additional ablation experiments in Tab.~\ref{tab:ablation_study_3}.  The results show that LossNet complements deep supervision and consistently improves segmentation performance, whereas applying a simple L2 loss directly at decoder scales does not yield comparable gains. This validates the effectiveness of the proposed feature-level perceptual supervision.}

\subsubsection{{Impact of Different Pre-trained Backbones in LossNet}}
{
We compare different pre-trained networks as the LossNet, as shown in Tab.~\ref{tab:ablation_study_lossnet_different_backbone}. The results show that both ResNet-50~\cite{ResNet} and Swin-T~\cite{Swin} perform worse than VGG-16~\cite{VGG} when used as perceptual supervision.
In fact, in perceptual loss based supervision, a relatively stable community consensus has been established, where most prior works~\cite{Ploss1,Ploss2,Esrgan} prefer to adopt VGG-16 or VGG-19 as pre-trained feature extractors. The primary reason is that VGG networks provide dense and well-structured hierarchical feature representations, which effectively capture low-level textures, mid-level structural patterns, and high-level semantic information. These multi-level perceptual cues are particularly suitable for segmentation tasks that require accurate modeling of boundary details and global structural consistency.
In comparison, ResNet introduces residual connections that enhance semantic abstraction but often lead to more mixed feature representations across layers. This characteristic may reduce the interpretability and stability of feature matching when used for perceptual supervision. Meanwhile, transformer-based architectures emphasize global context modeling through self-attention mechanisms. Although they provide strong semantic representation capability, their token-based feature representation is generally less sensitive to fine-grained spatial textures and boundary details, which are crucial for dense prediction tasks.
}

\begin{figure*}[t]
\centering
\includegraphics[width=\linewidth]{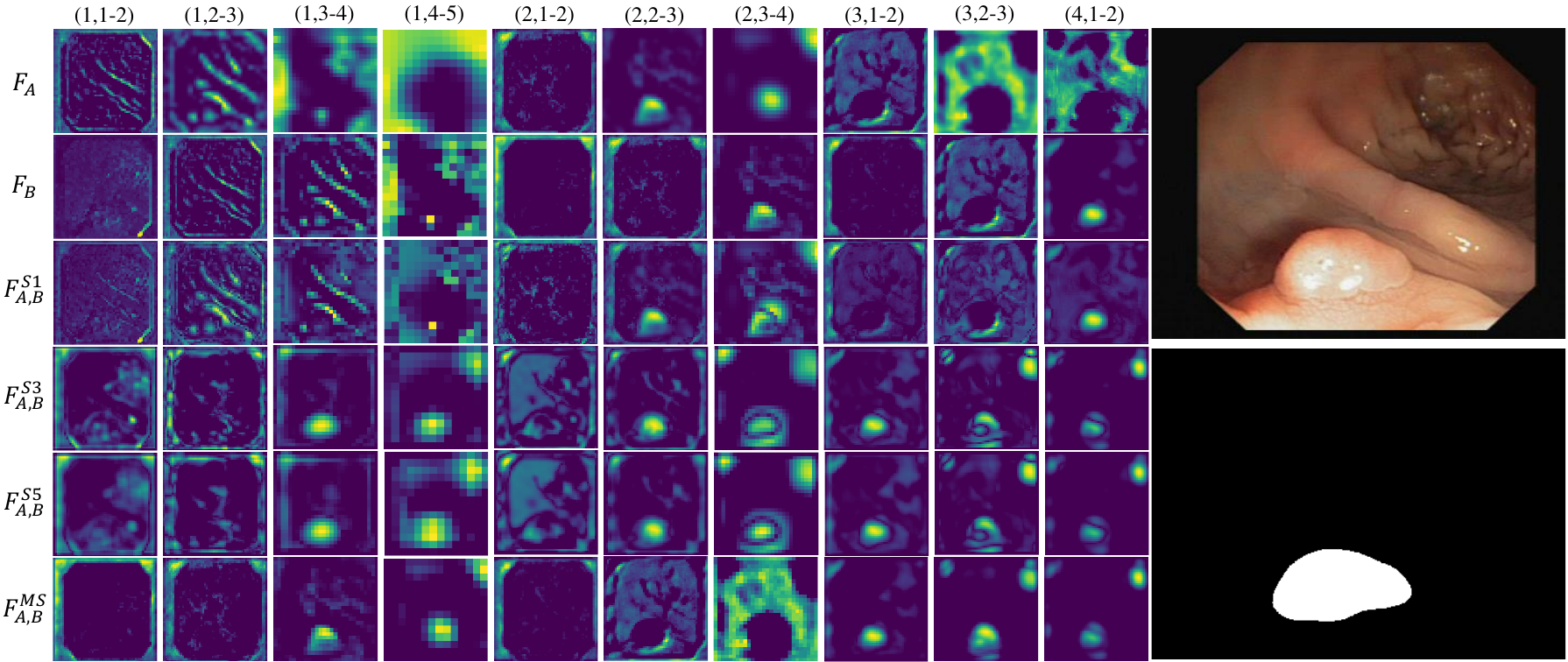}
\caption{Visualization of the feature maps in the multi-scale in multi-scale subtraction module.}
\label{fig:ms_module_all_feature_visual_1}
\end{figure*}

\subsubsection{Effectiveness of the intra-layer multi-scale subtraction design} 
Compared to the previous MSNet, the M$^{2}$SNet replace all the original single-scale subtraction unit with the stronger multi-scale subtraction unit. 
{As shown in Tab.~\ref{tab:ablation_study_4}, more scales of feature fusion can help improve the overall performance and multi-scale filters of size $[1, 3, 5]$ and $[1, 3, 5, 7]$ have close performance. 
{
This indicates that incorporating multi-scale difference modeling is beneficial, while excessively large receptive fields do not necessarily provide additional gains.
The selection of kernel sizes $1\times1$, $3\times3$, and $5\times5$ is motivated by balancing local detail preservation and moderate receptive field expansion. Specifically, the $1\times1$ kernel captures channel-wise interactions and preserves fine-grained spatial consistency. The $3\times3$ kernel models local structural patterns and boundary information, while the $5\times5$ kernel further enlarges the receptive field to encode mid-range contextual dependencies. These three scales jointly provide complementary local-to-mid-range spatial coverage without introducing excessive computational overhead.
We further investigate alternative kernel combinations, including $[1, 3, 7]$ and $[1, 5, 9]$, as reported in Tab.~\ref{tab:ablation_study_4}. The results show that larger-scale combinations do not yield consistent performance improvements compared with $[1, 3, 5]$, while increasing computational complexity. This suggests that deeper convolutional stacking already provides sufficient receptive field expansion, and overly large kernels may introduce redundant smoothing that weakens boundary modeling.
}
Next, we replace the fixed full one weights with Gaussian weights. It can be seen that Gaussian weights significantly degrade the performance of the model in all tasks. We think that the subtraction unit should try to maintain the characteristic distribution of the input features itself, but the Gaussian weights rigidly changes the spatial distribution of the original features, causing an extra burden for the subsequent decoder. Therefore, we choose the multi-scale convolution filters with fixed full one weights of size $[1, 3, 5]$ as the final setting.} 
To further show the advantages of our intra-layer multi-scale design, we apply other popular multi-scale modules (i.e., ASPP~\cite{ASPP} and DenseASPP~\cite{DenseASPP}) to the subtraction unit and these architectures are shown in Fig.~\ref{fig:aspp_denseaspp_ours_structure}. In Tab.~\ref{tab:ablation_study_5}, we thoroughly compare both the efficiency and accuracy of these three structures. It can be seen that ``M$^{2}$SNet (Ours)'' has a significant performance gain in terms of fourteen metrics on four challenges datasets under different tasks. However, the other two models not only increase the computational burden by more than 50\%, but also produce negative gains in multiple datasets. Therefore, the proposed intra-layer multi-scale subtraction design can be taken as a new baseline for future research in subtraction family.

To more intuitively show the differential information from different scales, we visualize all the features of the multi-scale in multi-scale subtraction module, as shown in Fig~\ref{fig:ms_module_all_feature_visual_1}. We can see that the multi-scale in multi-scale subtraction module can clearly highlight the difference between high-level features and other level features and propagate its localization effect to the low-level ones. At the same level, the intra-layer multi-scale aggregation design can comprehensively capture both the subtle and regional difference features. Thus, both the global structural information and local boundary information is well depicted in the enhanced features of different levels.

\begin{table*}[t]
  \centering
  \scriptsize
  \renewcommand{\arraystretch}{1.1}
  \setlength\tabcolsep{2pt}
    \caption{{Ablation experiments on different backbone architectures.}}\label{tab:ablation_study_backbone_1}
     \resizebox{\linewidth}{!}
     {
	\begin{tabular}{r|c||cccc|cccc|ccc|ccc}
	 \hline
	 \rowcolor{mygray}
  & &\multicolumn{4}{c|}{ETIS} & \multicolumn{4}{c|}{CVC-300-TV} & \multicolumn{3}{c|}{Lung Infection}& \multicolumn{3}{c}{BUSI}\\
	\rowcolor{mygray}
	 Methods&Backbone &mDice  $\uparrow$ & mIoU  $\uparrow$ & $F_\beta^w$  $\uparrow$&$E_\phi^{max}$  $\uparrow$& mDice  $\uparrow$ & mIoU  $\uparrow$ & $S_{\alpha}$ $\uparrow$&MAE$\downarrow$& DSC $\uparrow$ & $S_{\alpha}$ $\uparrow$&MAE$\downarrow$& Jaccard $\uparrow$&Specificity $\uparrow$ & Dice $\uparrow$\\
	\hline
M$^{2}$SNet&Res2Net-50~\cite{pami20Res2net}   & {0.749} &0.678 &{0.712} & {0.872}&{0.876} &{0.805} &{0.918} & {0.010}&{0.795} &{0.855} & {0.006}&71.52$\pm$1.19&98.35$\pm$0.20&79.21$\pm$1.38 \\
M$^{2}$SNet&ConvNeXt-B~\cite{ConvNeXt}   & {0.776} &0.703 &{0.748} & {0.896}&{0.894} &{0.839} &{0.942} & {0.008}&{0.817} &{0.868} & {0.005}&73.64$\pm$1.48&98.96$\pm$0.25&81.74$\pm$1.55 \\
M$^{2}$SNet&PVTv2-B4~\cite{PVTv2}   & {0.785} &0.710 &{0.754} & {0.901}&{0.899} &{0.843} &{0.945} & {0.008}&{0.815} &{0.863} & {0.005}&73.94$\pm$1.32&98.84$\pm$0.22&81.28$\pm$1.48 \\
	\hline
	
	\end{tabular}
	}
\end{table*}

 \subsubsection{{Effectiveness of using different backbones}} 
{
The motivation for using Res2Net-50 in this paper for extensive comparison is as follows:
(1) \textbf{Stronger Multi-scale Feature Representation.} 
Res2Net introduces hierarchical residual-like connections within each residual block by splitting feature channels into multiple groups. This design enables the backbone to capture multi-scale receptive fields inside a single block, allowing the model to simultaneously learn fine-grained local details and long-range contextual information. Such capability is particularly beneficial for medical image segmentation, where target structures often present large scale variations.
(2) \textbf{Better Compatibility with the Proposed Multi-scale Interaction Strategy.} 
Our framework relies on multi-scale in multi-scale subtraction operations to model complementary information across hierarchical feature levels. The enhanced intra-block multi-scale representations provided by Res2Net produce richer scale-aware features, which help the subtraction mechanism better highlight structural differences between feature levels, thereby improving boundary localization and structural consistency.
(3) \textbf{Fair Comparison and Methodological Continuity.}
The earlier conference version of this work focuses on polyp segmentation, where most state of the art polyp segmentation methods adopt Res2Net as the backbone. To ensure fair comparison with existing approaches and maintain methodological continuity between the conference version and the extended manuscript, we consistently adopt Res2Net-50 in this work.
}

{
We further conduct experiments using a stronger CNN-based backbone, ConvNeXt-B~\cite{ConvNeXt}, and a transformer-based backbone, PVTv2-B4~\cite{PVTv2}. As shown in Tab.~\ref{tab:ablation_study_backbone_1}, the segmentation performance is further improved with both backbones. These results demonstrate that the proposed M$^2$SNet is compatible with more advanced backbone architectures and exhibits strong generalization capability across different network paradigms.
}

\section{Discussion and Future Work}
{
\textbf{Subtraction Operation}:
We provide a theoretical interpretation to explain why subtraction can effectively reduce feature redundancy compared with addition and concatenation. (1) \textbf{Feature Fusion from Linear Signal Decomposition Perspective.} 
Let $F_l$ and $F_h$ denote two hierarchical feature representations extracted from adjacent network levels. These features can be decomposed into shared components and complementary components:
\begin{equation}
F_l = S + C_l, \quad F_h = S + C_h
\end{equation}
where $S$ represents shared semantic information and $C_l, C_h$ denote scale-specific complementary information. 
Under this formulation, conventional addition-based fusion can be expressed as:
\begin{equation}
F_{add} = F_l + F_h = 2S + C_l + C_h .
\end{equation}
This formulation preserves and amplifies the shared information $S$, which increases feature redundancy and may cause the fused representation to be dominated by duplicated semantic responses. 
Similarly, concatenation-based fusion can be written as:
\begin{equation}
F_{cat} = [F_l, F_h] .
\end{equation}
This operation retains all feature components, including duplicated shared signals. As a result, the network must rely on additional convolution layers to implicitly suppress redundancy, which increases parameter complexity and computational overhead. 
In contrast, subtraction-based fusion can be formulated as:
\begin{equation}
F_{sub} = F_h - F_l = C_h - C_l .
\end{equation}
This formulation explicitly cancels the shared component $S$, enabling subtraction to naturally suppress redundant information by isolating complementary structural signals between hierarchical features. 
(2) \textbf{Correlation Suppression and Orthogonalization Effect.} 
Hierarchical feature maps in encoder--decoder architectures are typically highly correlated due to progressive feature abstraction. From a statistical perspective, addition and concatenation preserve correlated feature components, while subtraction acts as a decorrelation operator. By suppressing correlated shared responses, subtraction encourages the network to learn representations that are closer to complementary or orthogonal feature subspaces, improving representation diversity and discriminability.
(3) \textbf{Frequency Domain Interpretation.}  
Hierarchical features often encode semantic information at different spatial frequency ranges. Lower-level features usually contain high-frequency structural details, while higher-level features capture low-frequency semantic context. Subtraction behaves similarly to a differential operator that enhances structural transitions and high-frequency residual components. This property helps emphasize boundary details and fine structures that are crucial for precise segmentation.
}
\\
\textbf{Multi-scale Subtraction Unit}:
Different from previous addition and concatenation operations, 
using subtraction in multi-level structure make resulted features input to the
decoder have much less redundancy among different levels and their level-specific
properties are significantly enhanced. In this work, we further explore the potential of the subtraction unit in intra-layer multi-scale fusion. How to improve the accuracy while maintaining the same efficiency as the single-scale one is the key challenge. We provide the solution of using multi-scale convolution filters with fixed parameters. Compared to the single-scale design, multi-scale subtraction unit can enable the network to collect more complementary information both in pixel-pixel and neighbor-neighbor levels. The advantages of multi-scale subtraction unit in terms of efficiency and accuracy can be seen in Tab.~\ref{tab:ablation_study_5}. Multi-scale information extraction and feature aggregation are two general problems in the field of computer vision. Our multi-scale subtraction unit can solve both of them at once. We think this new paradigm can drive more researches on the subtraction operation in the future.
\\
\textbf{LossNet}: LossNet is similar in form to perception loss~\cite{Ploss} that has been applied in many tasks, such as style transfer and inpainting.  While in those vision tasks, the perception-like loss is mainly used to speed the convergence of GAN and obtain high frequency information and ease checkerboard artifacts, but it does not bring obvious accuracy improvement. In our paper, the inputs are binary segmentation masks, LossNet can directly target the geometric features of the lesion and perform joint supervisions from the contour to the body, thereby improving the overall segmentation accuracy.
\\
{
\textbf{Limitation}: Despite these advantages, several limitations remain. First, the current framework is fully supervised and relies on sufficient annotated training data. In extremely low-data or unseen clinical scenarios, the generalization ability may still be constrained. Second, the multi-scale subtraction unit adopts fixed kernel configurations, which, although computationally efficient and empirically effective, may not fully capture very long-range dependencies compared to dynamic convolution or attention-based mechanisms. Third, while LossNet enhances representation-level supervision, it introduces additional computational overhead during training, which may limit scalability for very large models or high-resolution volumetric inputs.
\\
\textbf{Future Work}: Future research can be explored in several directions. One promising direction is to design adaptive or dynamic multi-scale subtraction mechanisms that can automatically adjust receptive fields according to lesion size and modality characteristics. Another direction is to investigate lightweight perceptual supervision strategies or self-supervised representation alignment methods to reduce training cost while maintaining structural consistency. Furthermore, integrating foundation-model-based pretraining or prompt-driven adaptation may improve generalization under limited annotations. Finally, extending the proposed framework to 3D volumetric segmentation and multi-modal medical imaging scenarios could further enhance its practical applicability in real clinical settings.
}

\section{Conclusion}
In this paper, we rethink previous addition-based or concatenation-based methods and present a simple yet general multi-scale in multi-scale subtraction network (M$^{2}$SNet) for more efficient medical image segmentation.   
Based on the proposed intra-layer multi-scale subtraction unit, we pyramidally aggregate adjacent levels to extract lower-order and higher-order cross-level complementary information and combine with level-specific information to enhance multi-scale feature representation. 
Besides, we design a loss function based on a training-free network to supervise the prediction from different feature levels, which can optimize the segmentation on both structure and details during the backward phase.  
Experimental results on $11$ benchmark datasets towards $4$ medical segmentation tasks demonstrate that the proposed model outperforms various state-of-the-art methods.

\section*{Acknowledgments}
 This work was supported by Dalian Science and Technology Innovation Foundation under Grant 2023JJ12GX015, and by the National Natural
 Science Foundation of China under Grant 62276046 and 62431004.  (Corresponding author: Lihe Zhang.)

\section*{Conflicts of Interests}
The authors declared that they have no conflicts of interest to this work.

\bibliographystyle{unsrt}
\bibliography{mir-article}

\newpage

\begin{figure}[htbp]%
\centering
\includegraphics[width=0.22\textwidth]{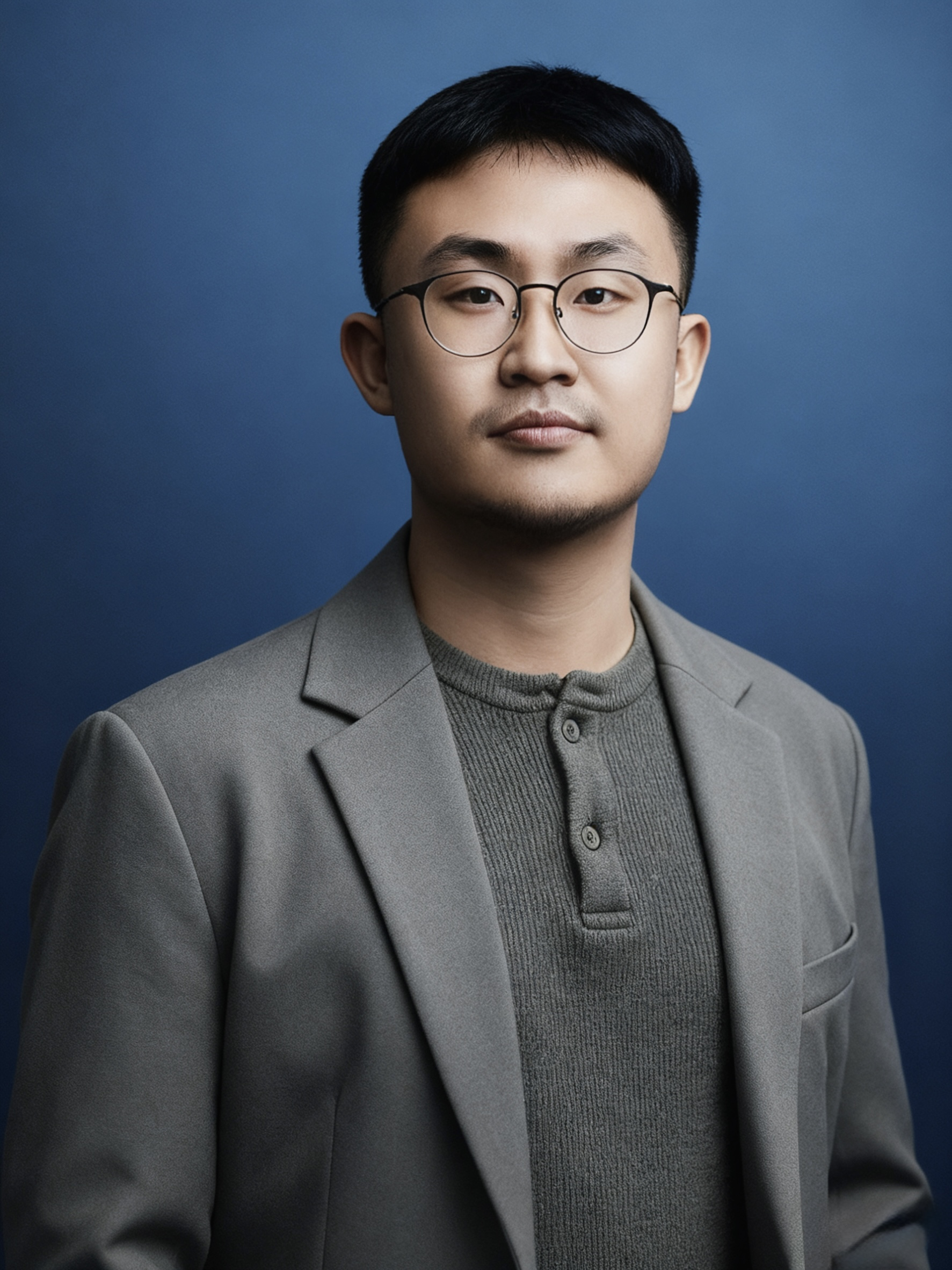}
\end{figure}
\noindent{\bf Xiaoqi Zhao }\quad
received the Ph.D. degree from Dalian University of Technology in 2024. He is an AI4X Principal Investigator / Postdoctoral Fellow at Nanyang Technological University (NTU) in Singapore, supported by both the NRF Postdoctoral Award and the NTU AI-for-X Postdoctoral Fellowship. 
He was a postdoctoral researcher at Yale University in 2025. His research interests include self-driven learning mode, unified visual perception, AI4Industry and AI4Health. He was nominated for the 2025 WAIC Yunfan Award, recognized as an outstanding reviewer at CVPR 2022, and has presented oral/highlight papers at CVPR, ECCV, and ACM MM. E-mail: zhaoxq.cv@gmail.com,  ORCID iD: 0000-0001-7734-5128

\begin{figure}[htbp]%
\centering
\includegraphics[width=0.22\textwidth]{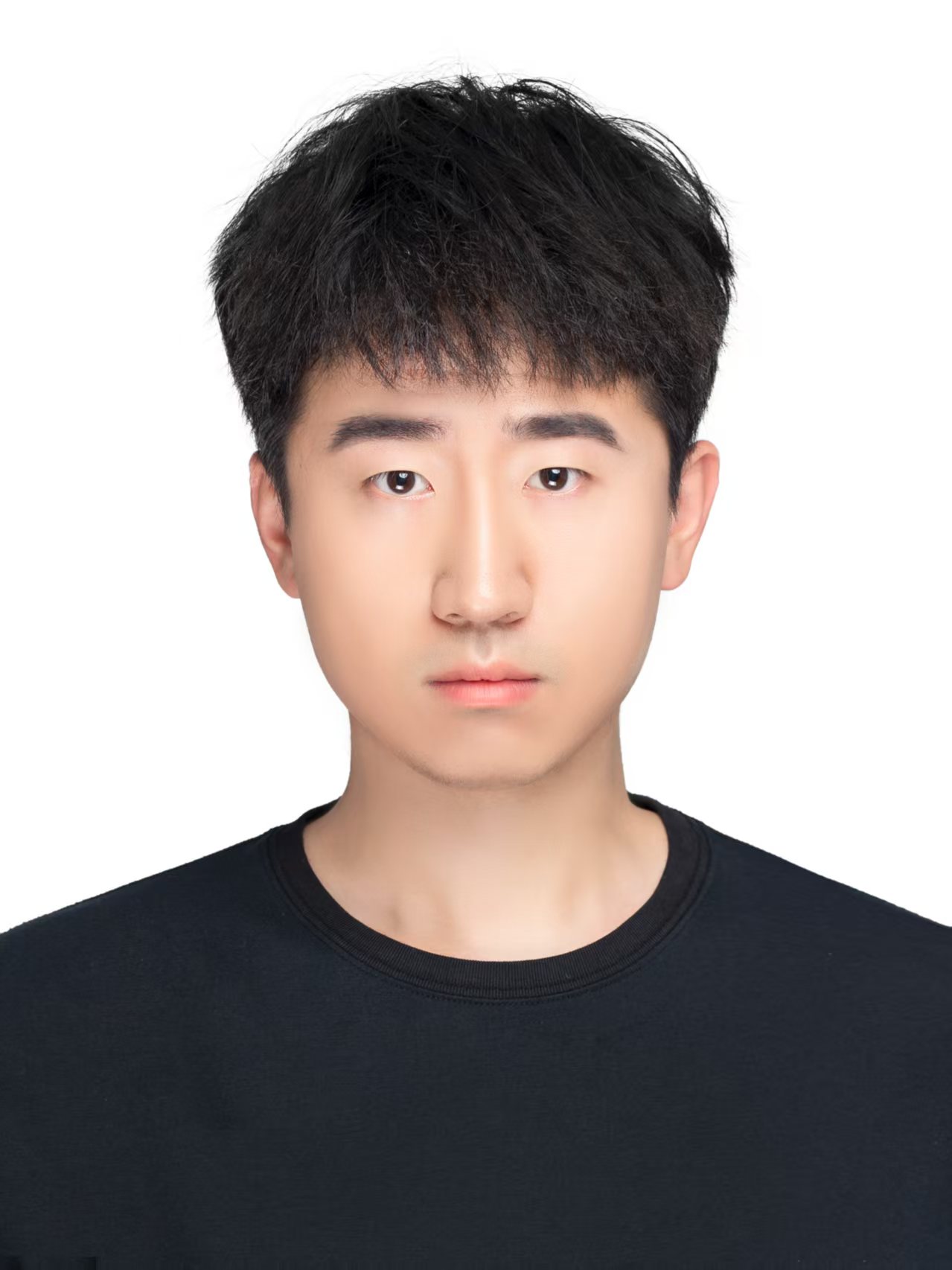}
\end{figure}
\noindent{\bf Hongpeng Jia}\quad
received his Master’s degree in Information and Communication Engineering from Dalian University of Technology. During his postgraduate studies, his research focused on computer vision and AIGC-related technologies in the field of artificial intelligence. He was awarded the National Scholarship and honored as an Outstanding Graduate of Liaoning Province. Currently, he works in the Digitalization Team of Shenyang Cigarette Factory, engaged in the construction and application of enterprise information systems. E-mail: a2916956058@163.com

\begin{figure}[htbp]%
\centering
\includegraphics[width=0.22\textwidth]{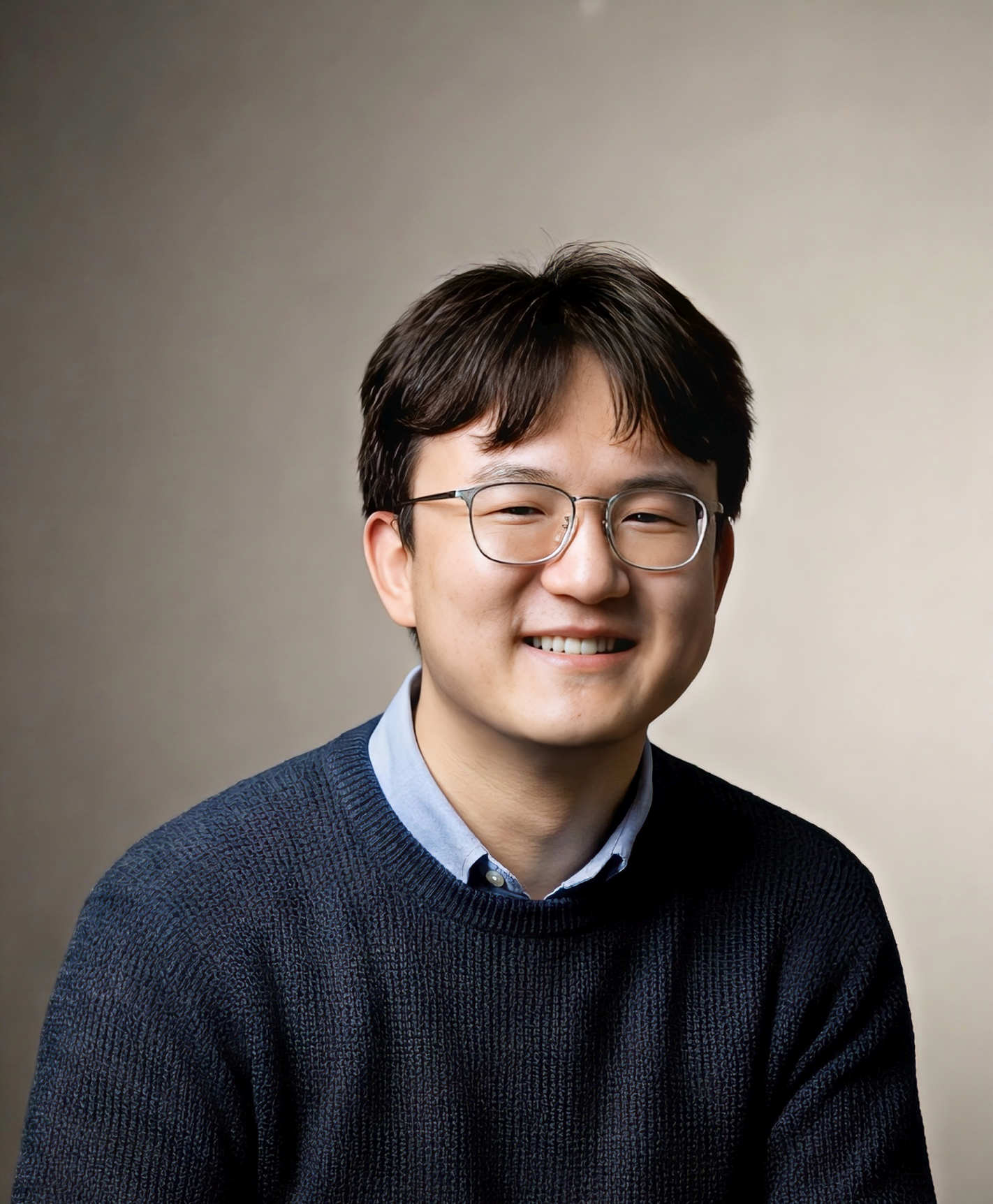}
\end{figure}
\noindent{\bf Youwei Pang }\quad
received the Ph.D.~degree in Signal and Information Processing from the School of Information and Communication Engineering, Dalian University of Technology (DUT), China, in 2025. He joined Nanyang Technological University as a postdoctoral researcher in 2025. His research interests include multi-task learning, multi-modal learning, efficient architecture design, and industrial AI. He has published  about 40 top journal and conference papers such as TPAMI, IJCV, CVPR, ICCV, etc. E-mail: lartpang@gmail.com

\begin{figure}[htbp]%
\centering
\includegraphics[width=0.22\textwidth]{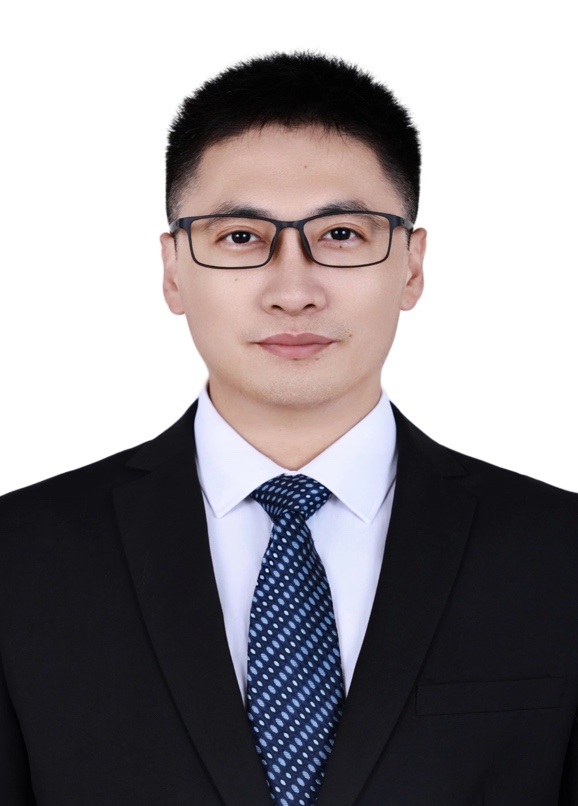}
\end{figure}
\noindent{\bf Long Lv }\quad
received Master of Medicine and graduated from the Seven-Year Program in Clinical Medicine of Dalian Medical University in 2014. He is currently an Associate Chief Physician at Zhongshan Hospital Affiliated of Dalian University. His research interests include urinary system tumors, medical data analysis and artificial intelligence.  E-mail:  lvlong113@126.com

\begin{figure}[htbp]%
\centering
\includegraphics[width=0.22\textwidth]{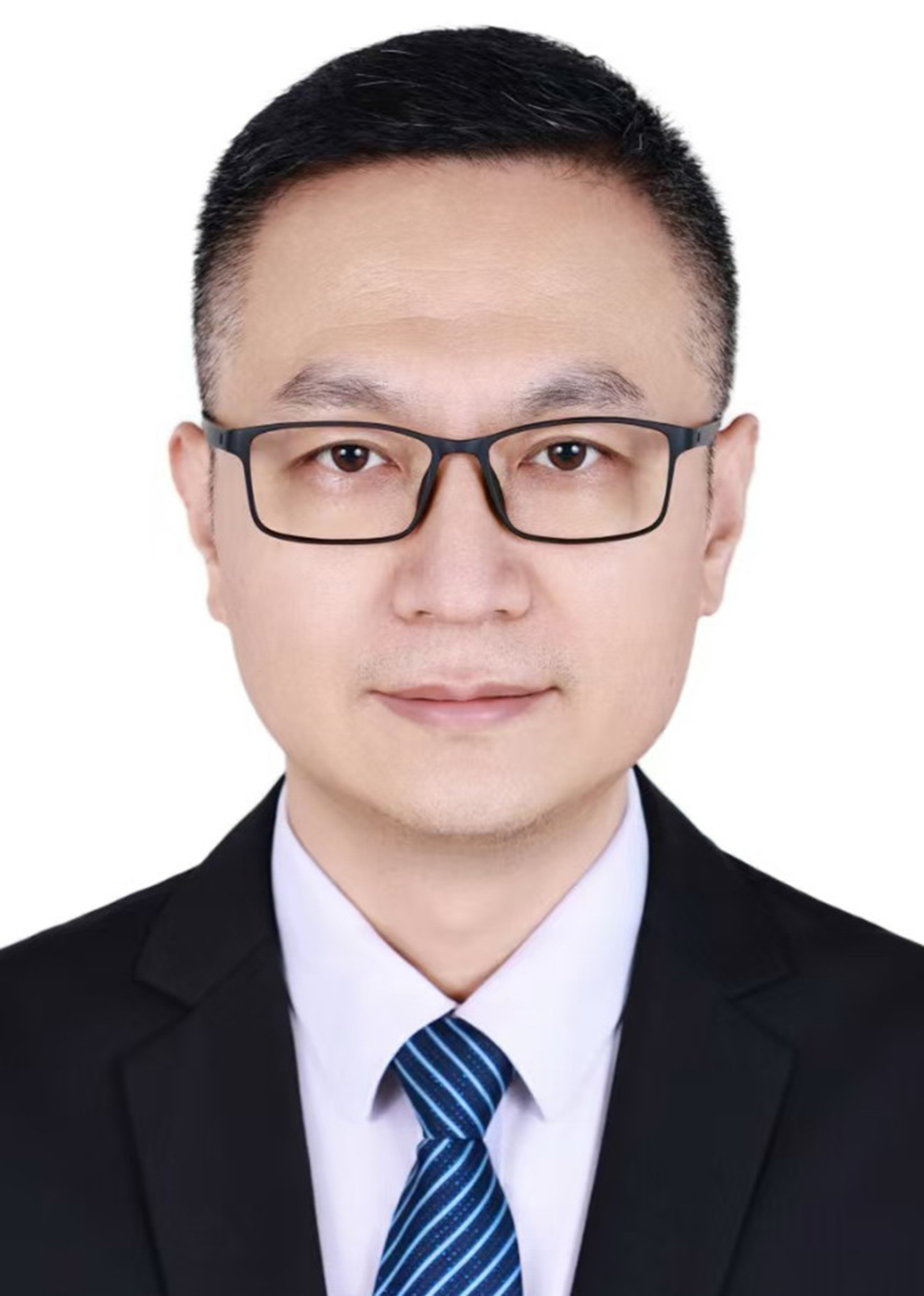}
\end{figure}
\noindent{\bf Feng Tian }\quad
received his MD degree in the Fourth Military Medical University (FMMU), Xi’an, China, in 2008. He is the Deputy Director of the Key Laboratory of Microenvironment Regulation and Immunotherapy of Urinary Tumors of Liaoning Province and Chief Physician of Urology of the Affiliated Zhongshan Hospital of Dalian University. His current research interests include composition analysis of urinary stone and AI recognition and analysis of CT image of urinary stone.   E-mail:  tianfeng\_dl@sohu.com

\begin{figure}[htbp]%
\centering
\includegraphics[width=0.22\textwidth]{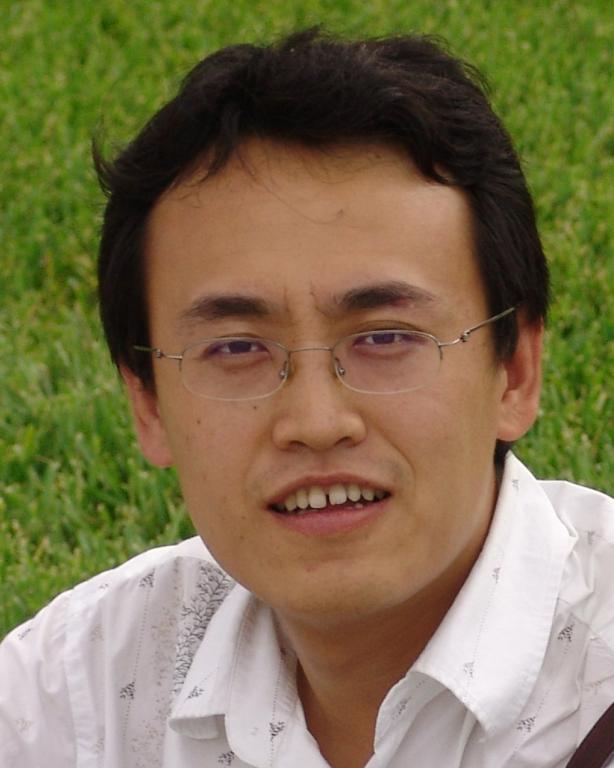}
\end{figure}

\noindent{\bf Lihe Zhang }\quad (Member, IEEE) received the M.S. degree in signal and information processing from Harbin Engineering University (HEU), Harbin, China, in 2001, and the Ph.D. degree in signal and information processing from the Beijing University of Posts and Telecommunications (BUPT), Beijing, China, in 2004. He is currently a Full Professor with the School of Information and Communication Engineering, Dalian University of Technology (DUT), Dalian, China.  His current research interests include computer vision and pattern recognition. 
E-mail: zhanglihe@dlut.edu.cn (Corresponding author), ORCID iD: 0000-0003-4648-4437

\begin{figure}[htbp]%
\centering
\includegraphics[width=0.20\textwidth]{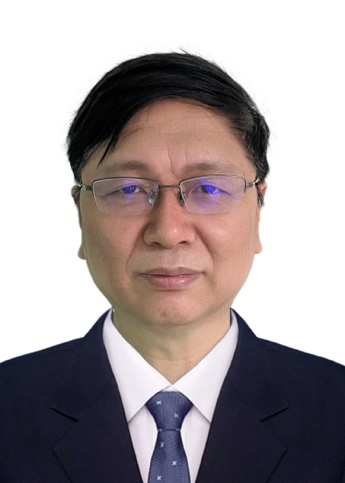}
\end{figure}
\noindent{\bf Weibing Sun }\quad
is currently Director of the Department of Urology at Zhongshan Hospital Affiliated to Dalian University, and Director of the Key Laboratory of Microenvironment Regulation and Immunotherapy of Urinary Tumors in Liaoning Province. He received his bachelor degree from China Medical University, and his master degree in Surgery and PhD in Integrated Traditional Chinese and Western Medicine from Dalian Medical University. His research focuses on clinical and basic research of urinary system tumors.   E-mail:  massurm@163.com

\begin{figure}[htbp]%
\centering
\includegraphics[width=0.20\textwidth]{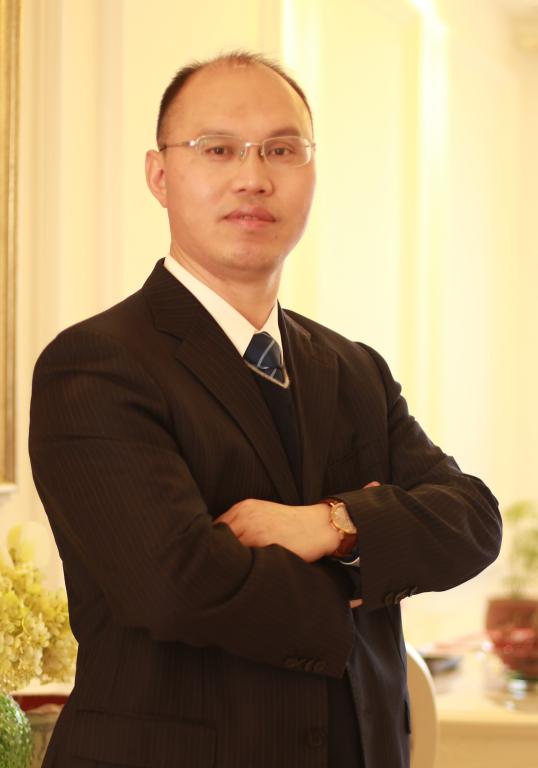}
\end{figure}

\noindent{\bf Huchuan Lu }\quad  (Fellow, IEEE) received the M.S. degree in signal and information processing and the PhD. degree in system engineering from the Dalian University of Technology (DUT), Dalian, China, in 1998 and 2008, respectively.  He joined the faculty in 1998 and currently is a full professor with the School of Information and Communication Engineering, DUT. His current research interests include computer vision and pattern recognition with focus on visual object tracking, saliency detection, and segmentation. 
E-mail: lhchuan@dlut.edu.cn

\end{document}